\documentclass[letterpaper]{article} 
\usepackage[preprint]{aaai2027}  
\usepackage[hyphens]{url}  
\usepackage{graphicx} 
\usepackage{natbib}  
\usepackage{caption} 
\usepackage{algorithm}
\usepackage{algorithmic}
\usepackage{multirow}
\usepackage{subcaption}
\usepackage{amssymb}
\usepackage{amsmath}
\usepackage{pifont}
\usepackage{makecell}
\usepackage{colortbl}
\usepackage{xcolor}     
\usepackage{colortbl}   

\usepackage{soul}
\usepackage{newfloat}
\usepackage{listings}
\DeclareCaptionStyle{ruled}{labelfont=normalfont,labelsep=colon,strut=off} 
\floatstyle{ruled}
\newfloat{listing}{tb}{lst}{}
\floatname{listing}{Listing}

\usepackage{booktabs}
\usepackage{xcolor}     
\usepackage{colortbl}   

\title{Dual-Prototype Disentanglement: A Context-Aware \\Enhancement Framework for Time Series Forecasting}
\author{
    Haonan Yang\textsuperscript{\rm 1}\equalcontrib,
    Yulin He\textsuperscript{\rm 1}\equalcontrib,
    Jianchao Tang\textsuperscript{\rm 1}\corresponding,
    Zhuo Li\textsuperscript{\rm 1},
}
\affiliations{
    \textsuperscript{\rm 1}College of Computer Science and Technology, National University of Defense Technology,

    Changsha 410073, China\\
    \{yanghaonan21, heyulin13, tangjianchao14, lizhuo19\}@nudt.edu.cn
}

\begin{document}

\maketitle

\begin{abstract}
Real-world time series are governed by both recurring structures, such as trends and seasonality, and infrequent yet critical variations, such as abrupt shifts and rare events. However, existing methods often lack an explicit mechanism to organize and utilize these heterogeneous patterns according to their distinct forecasting roles. Consequently, common and rare patterns can become entangled, preventing models from dynamically distinguishing and selectively leveraging them according to context. To address this issue, we propose Dual-Prototype Adaptive Disentanglement (DPAD), a model-agnostic framework that organizes temporal patterns by their forecasting roles. Specifically, we construct a Dynamic Dual-Prototype bank (DDP), comprising a common pattern bank initialized with structured temporal priors to represent prevalent dynamics, and a rare bank that adaptively memorizes infrequent deviations. Then a Dual-Path Context-aware routing (DPC) mechanism enhances outputs with selectively retrieved context-specific pattern representations from DDP. A Disentanglement-Guided Loss (DGLoss) is further introduced to ensure that each prototype bank specializes in its designated role while maintaining sufficient coverage. Extensive experiments across diverse real-world benchmarks demonstrate that DPAD consistently improves the forecasting performance of a range of time-series models. Code is available in supplementary materials.
\end{abstract}


\section{Introduction}\label{Introduction}

Time series forecasting (TSF) plays a pivotal role in numerous real-world domains, including weather forecasting \cite{Kaifeng2023,Haixu2023}, transportation scheduling \cite{Chenjuan2020,Jin2021} and energy consumption \cite{Alvarez2011,Guo2015}. Despite remarkable progress in deep forecasting models \cite{TimeMixer2024,PatchTST2023}, time series remain challenging due to their inherent complexity \cite{TSL2024, chi2021t, qiu2025}. Prevalent recurring modes, such as trends and seasonalities, often coexist with infrequent but prediction-relevant deviations, abrupt shifts, and local fluctuations. 

Recent studies address this challenge mainly through architectural modifications \cite{TimesNet2023,Autoformer2021} or enhancement strategies \cite{Kim2022,Zhiding2023}. To effectively capture complex temporal dependencies, some approaches refine attention mechanisms or redesign MLP and CNN architectures to better model long-range and cross-channel dependencies \cite{tang2024,moderntcn2024}. In parallel, enhancement strategies introduce additional temporal knowledge without substantially modifying the backbone. Retrieval-augmented methods retrieve similar historical samples to supplement the current context, post-hoc revision methods correct prediction errors or distributional discrepancies after forecasting \cite{PIR2025}, and specialized objectives provide structural or component-wise supervision during training \cite{FreDF2025}. 


\begin{figure*}[!t]
  \centering
    \begin{subfigure}[t]{0.3\textwidth}
        \includegraphics[width=\textwidth]{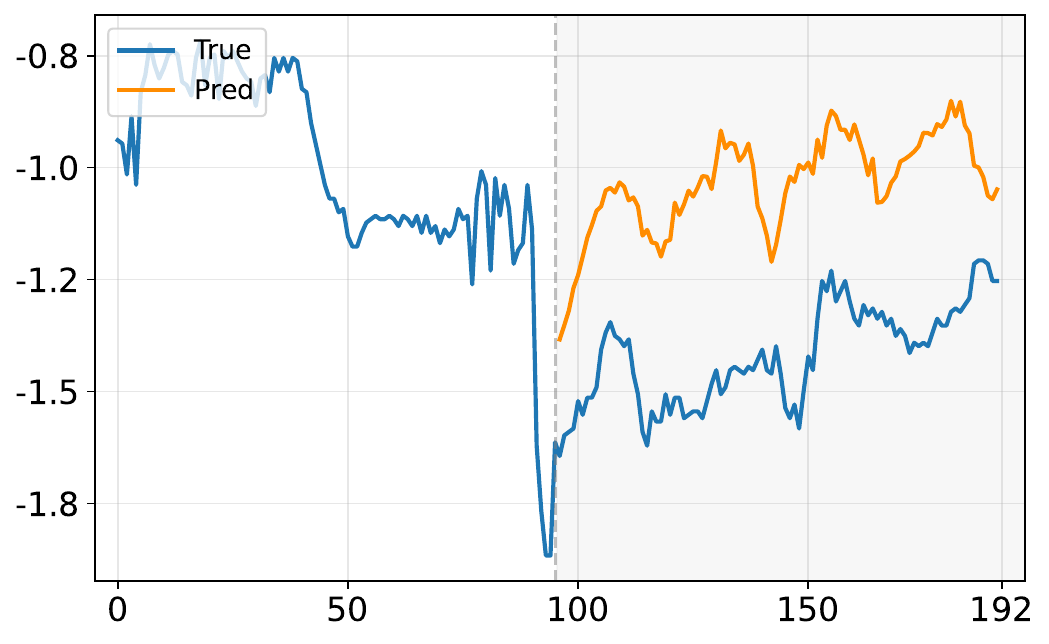}
        \caption{Failure under Distribution Shifts}
        \label{introd_a}
    \end{subfigure}
    \hspace{-0.05cm}
    \begin{subfigure}[t]{0.3\textwidth}
        \includegraphics[width=\textwidth]{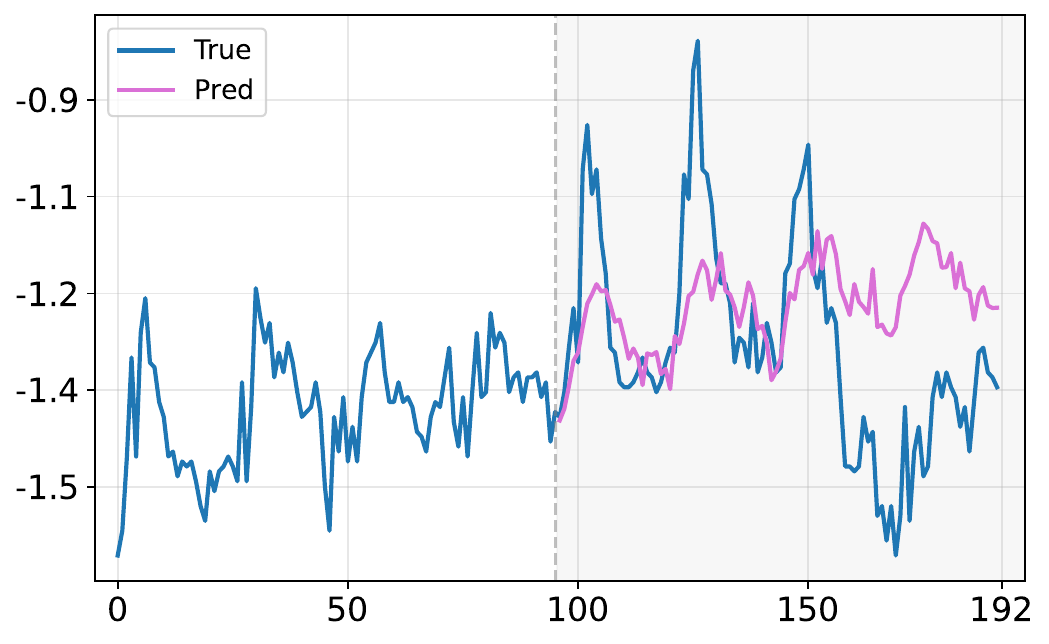}
        \caption{Failure on Intertwined Patterns}
        \label{introd_b}
    \end{subfigure}
    \hspace{-0.05cm}
    \begin{subfigure}[t]{0.3\textwidth}
        \includegraphics[width=\textwidth]{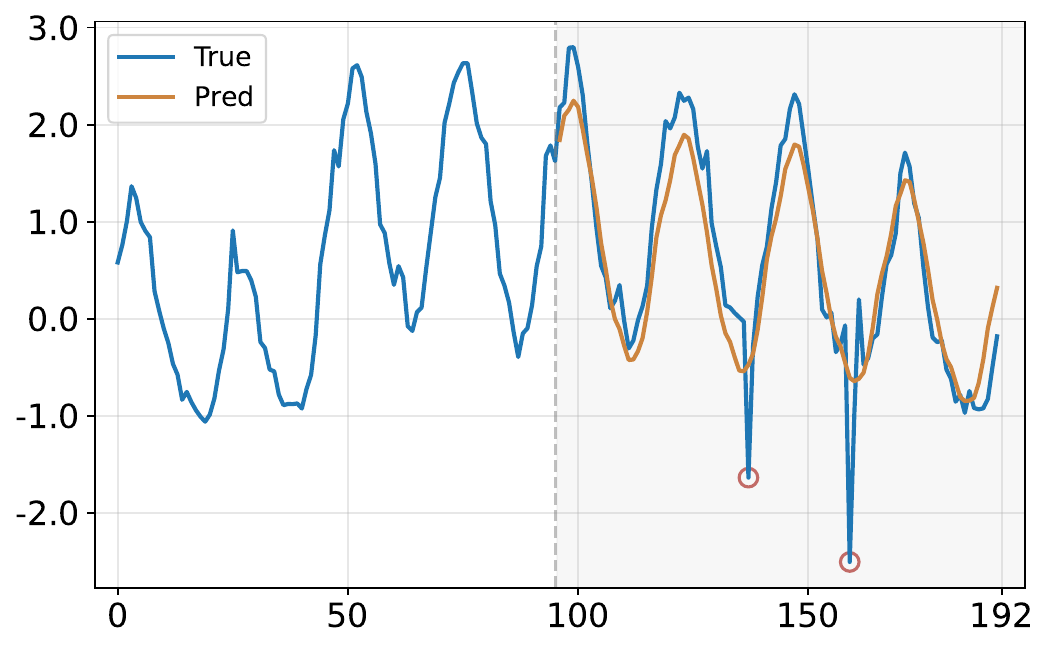}
        \caption{Failure without Rare Events}
        \label{introd_c}
    \end{subfigure}
    \caption{Motivating examples of challenging forecasting scenarios. We illustrate three cases where a vanilla forecasting backbone may exhibit large errors: (a) abrupt distribution shifts, (b) intertwined complex patterns, and (c) critical rare events.}
    \label{Introd}
\end{figure*}

Nevertheless, existing methods primarily enhance pattern modeling or correct forecasting outcomes, without explicitly organizing and utilizing heterogeneous patterns by forecasting roles \cite{Qiuz2024, yang2025}. In real-world time series, prevalent recurring patterns and infrequent yet critical deviations contribute differently in forecasting \cite{han2025, TimeMixer2025}. Although such patterns have been recognized in prior studies, they are rarely explicitly organized and utilized through role-specialized memory and context-aware routing. Without role-specific organization, these patterns can become entangled, while existing enhancement methods like uniform enhancement or similarity-based retrieval further limits their context-aware selection \cite{AMD2025, kudrat2025}. Consequently, models may struggle to dynamically distinguish and selectively leverage common and rare patterns according to the input context. As shown in Figure \ref{Introd}, this limitation can lead to forecasting errors under abrupt distribution shifts, intertwined temporal structures, and critical rare events. Therefore, the key issue is not merely how to model more temporal patterns, but how to organize them by forecasting roles and selectively utilize them according to context \cite{ning2025}.


To address these limitations, we propose the Dual-Prototype Adaptive Disentanglement framework (DPAD), a model-agnostic enhancement framework that organizes temporal patterns into a disentangled memory according to their forecasting roles. Specifically, DPAD maintains two role-specialized prototype banks in the Dynamic Dual-Prototype bank (DDP), which comprises a common pattern bank covering prevalent modes, and a rare pattern bank dedicated to memorizing infrequent yet critical events. Then, a Dual-Path Context-aware routing (DPC) mechanism dynamically retrieves and activates relevant patterns from the DDP, thereby enhancing the backbone's representation via a weighted gating strategy. Moreover, a Disentanglement-Guided Loss (DGLoss) jointly promotes separation between common and rare patterns, enforces diversity among common prototypes, and maintains the distinctiveness of rare prototypes. Our contributions are summarized as follows:


\begin{itemize}
    \item We identify role-agnostic pattern organization as a key limitation in time series forecasting and address it by disentangling heterogeneous temporal patterns according to their forecasting roles and selectively leveraging them based on the input context.
  \item We propose DPAD, a model-agnostic auxiliary framework that employs a learnable dual-prototype bank and an adaptive routing mechanism to dynamically enhance the model's forecasting performance, and further introduces a disentanglement-guided loss that enforces effective pattern separation and preservation while jointly optimizing the backbone and the prototype banks.
  \item Extensive experiments across diverse real-world datasets and state-of-the-art backbones show that DPAD consistently enhances forecasting performance with modest absolute overhead for most evaluated backbones.
\end{itemize}

\section{Related Work}

\subsection{Time Series Forecasting Models}
Deep forecasting models have progressed significantly by learning expressive temporal representations from history \cite{Zezhi2025,qiu2025c}. Existing models can be broadly categorized into MLP-based, CNN-based, Transformer-based, and LLM-based approaches \cite{TSL2024}. MLP-based models \cite{NBEATS2019,FreTS2023} balance efficiency and performance through decomposition strategies and simple linear layers. CNN-based architectures \cite{TCN2018,SCINet2022,MiCN2023} leverage hierarchical convolutions and frequency-domain analysis to model local patterns and multi-periodicity. Transformer-based models \cite{Informer2021,pathformer2024} improve long-range dependency modeling through attention mechanisms and their variants, enabling effective modeling of global dependencies and cross-variable interactions. More recently, LLM-based approaches \cite{MOIARI2024,TimeMoE2025} adapt pre-trained LLMs to time series, exploring their generalization and in-context learning capabilities \cite{Timer2024}. These methods primarily improve the capacity of forecasting architectures. However, modeling more patterns does not necessarily enable a model to organize and utilize them according to their forecasting roles. When common and rare patterns coexist or become intertwined, increasing backbone capacity alone may be insufficient to dynamically distinguish and selectively leverage them under different contexts \cite{Ailing2023}. DPAD is orthogonal to this line of work: instead of modifying core architecture, DPAD serves as a model-agnostic auxiliary framework that provides role-specialized pattern organization and context-aware enhancement for different backbones.

\begin{figure*}[!t]
  \centering
    \includegraphics[width=0.9\textwidth]{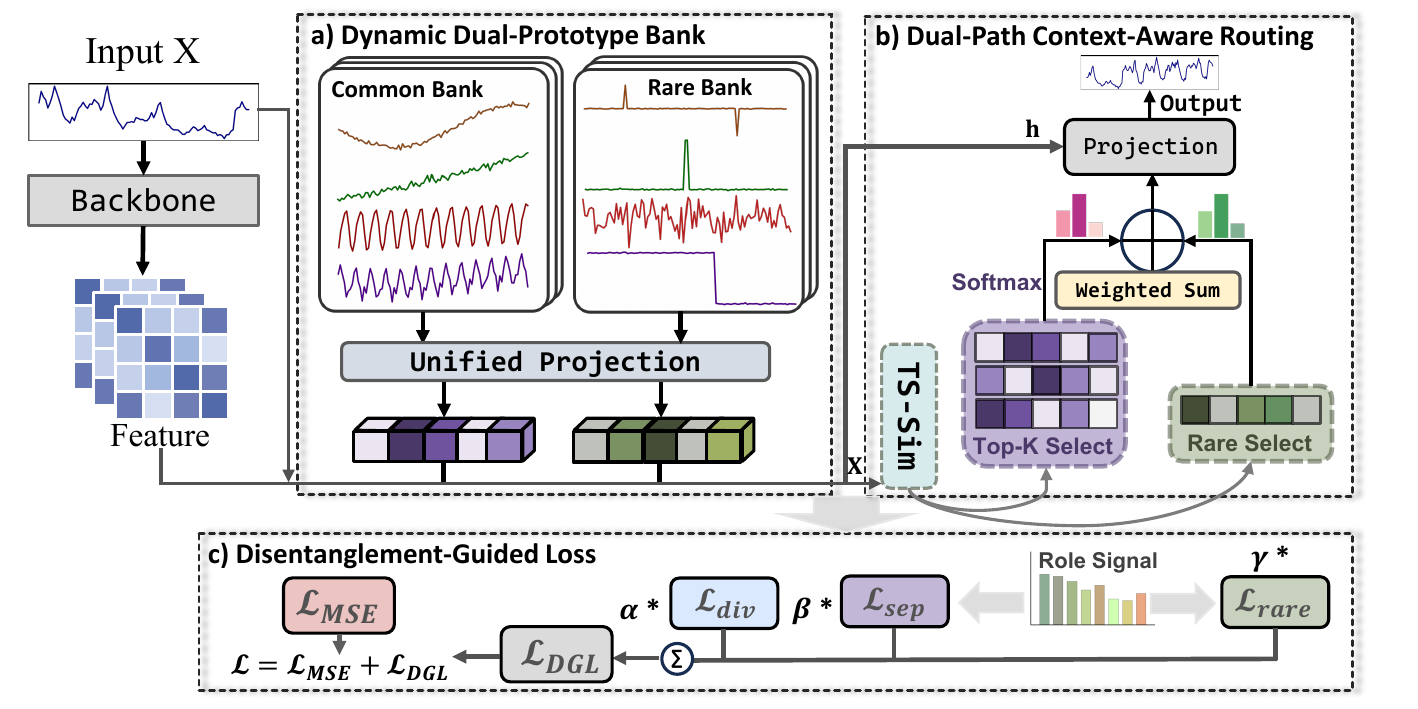}
    \caption{
      The overall architecture of the DPAD framework. The framework constructs a learnable dual-prototype bank, then performs adaptive dual-path routing on the input data to retrieve relevant prototypes, and finally generates a context-aware enhancement for prediction. The entire framework is optimized under a disentanglement-guided loss.
    }
    \label{pipeline}
\end{figure*}

\subsection{Enhance-Based Time Series Forecasting Methods}
Beyond architectural innovations, others introduce model-agnostic enhancement strategies. Retrieval-augmented methods retrieve relevant historical samples to provide additional context for prediction \cite{han2025, saraf2026}, but they typically require maintaining a historical sample database and performing sample-level similarity search at inference time. Post-hoc revision methods refine outputs after prediction by correcting instance-level errors or mitigating distribution shifts \cite{sun2024, uec2026}. They revise predictions after the backbone has produced outputs but do not directly support role-aware pattern organization. Specialized loss functions introduce structural or component-wise training objectives, such as decomposition-based supervision \cite{qiu2025} and patch-level statistical alignment \cite{FreDF2025}. However, they typically apply global constraints without introducing explicit pattern memory or context-aware retrieval mechanism. In summary, existing enhancement methods mainly improve forecasting from sample-level, output-level, or objective-level perspectives, while lacking an unified mechanism for organizing temporal patterns according to their forecasting roles and selectively utilizing them based on the input context. In contrast, DPAD constructs role-specialized prototype banks and dynamically retrieves context-specific common and rare patterns, avoiding the storage and online search of historical samples. This design enables role-aware enhancement of different patterns and context-aware enhancement with modest additional computational overhead.

\section{Method}
Time series forecasting aims to predict future values ${Y} = \{y_{L+1}, ..., y_{L+H}\}\in\mathbb{R}^{H\times{C}}$ from historical observations ${X} = \{x_1, ..., x_L\}\in\mathbb{R}^{L\times{C}}$, where $H$ is prediction horizon, $L$ is history horizon, and $C$ is number of variables.

\subsection{Structure Overview}
As illustrated in Figure \ref{pipeline}, DPAD consists of three key components: (a) Dynamic Dual-Prototype Bank (DDP) maintains common and rare pattern prototypes to explicitly disentangle different patterns according to their forecasting roles; (b) Dual-Path Context-aware routing (DPC) dynamically retrieves and activates relevant prototypes based on the input context, and then fuses the retrieved pattern representations with the backbone representation for enhanced prediction; (c) Disentanglement-Guided Loss (DGLoss) jointly optimizes the backbone and DDP by encouraging prototype specialization, separation, and diversity.

\subsection{Dynamic Dual-Prototype Bank}
The Dynamic Dual-Prototype bank (DDP) is designed to explicitly disentangle and memorize two distinct categories of temporal patterns: prevalent common patterns and critical yet infrequent rare patterns. Formally, DDP contains two separate prototype sets: a Common Pattern Bank $\mathcal{B}_c$ for stable temporal modes such as trends and seasonality, and a Rare Pattern Bank $\mathcal{B}_r$ for infrequent deviations, abrupt variations, and other prediction-relevant irregular patterns.

Specifically, we generate two groups of base sequences $\{\mathbf{s}_c^{i}\} \in \mathbb{R}^{{L}}$ and rare sequences $\{\mathbf{s}_r^{j}\} \in \mathbb{R}^{{L}}$ using distinct prior strategies to reflect different pattern characteristics:
\begin{equation}
\mathbf{s}_c^{i} \sim \mathcal{GP}\left(0, \, \lambda_{\text{l}} K_{\text{l}} + \lambda_{\text{r}} K_{\text{r}} + \lambda_{\text{p}} K_{\text{p}}\right),\quad i=1,\dots,M,
\end{equation}
\begin{equation}
\mathbf{s}_r^{j} \sim \mathcal{N}(0, \sigma^2 \mathbf{I}),\quad j=1,\dots,N,
\end{equation}
where $\mathbf{s}_c^{i}$ is the common sequence generated by a mixture of Gaussian Process ($\mathcal{GP}$) kernels to embed structured temporal priors, $K_{\text{l}}, K_{\text{r}}, K_{\text{p}}$ are the Linear, radial basis function (RBF), and Periodic kernel, and $\lambda_i$ are mixing coefficients, all shared across datasets. This yields diverse yet structured prototypes for stable and recurring patterns, thereby providing a well-initialized basis for modeling prevalent temporal dynamics.

In contrast, $\mathbf{s}_r^{j}$ is initialized from a small-variance Gaussian distribution $\mathcal{N}(0, \sigma^2 \mathbf{I})$. This initialization provides a weakly biased starting point, allowing rare prototypes to remain flexible and adaptively memorize infrequent deviations during training. After initialization, all prototypes are treated as learnable parameters and optimized jointly with backbone through backpropagation at every training iteration.

Then each generated sequence is projected into a shared $D$-dimensional latent space via the linear layers:
\begin{equation}
\mathbf{p}_c^{i} = \mathbf{Proj}_c(\mathbf{s}_c^{i}),
\end{equation}
\begin{equation}
\mathbf{p}_r^{j} = \mathbf{Proj}_r(\mathbf{s}_r^{j}),
\end{equation}

The resulting Common Pattern Bank $\mathcal{B}_c$ and Rare Pattern Bank $\mathcal{B}_r$ are defined as follows:
\begin{equation}
\mathcal{B}_c = \{\mathbf{p}_c^{1}, \mathbf{p}_c^{2},\dots,\mathbf{p}_c^{M}\},\quad \mathcal{B}_r = \{\mathbf{p}_r^{1}, \mathbf{p}_r^{2},\dots,\mathbf{p}_r^{N}\},
\end{equation}
This strategy explicitly organizes common and rare patterns according to their forecasting roles while keeping the prototype memory compact and adaptive. Both $\mathcal{B}_c$ and $\mathcal{B}_r$ are jointly optimized with the backbone under DGLoss. Consequently, the common prototypes evolve toward representative and dataset-specific prevalent dynamics, while the rare prototypes adapt to infrequent yet prediction-relevant variations.

\subsection{Dual-Path Context-Aware Routing}
The Dual-Path Context-aware routing (DPC) mechanism is proposed to adaptively enhance final prediction. Given input $\mathbf{X}$ and backbone representation $\mathbf{h} \in \mathbb{R}^{B \times C \times D}$, DPC uses $\mathbf{X}$ to dynamically retrieve context-relevant prototypes from DDP and fuses the retrieved representations with $\mathbf{h}$ for prediction. Specifically, we first use $\phi(\cdot)$ to project $\mathbf{X}$ into the channel-wise temporal representation of shape $\mathbb{R}^{B\times C\times D}$ to match the prototypes. Then TS-Sim computes the similarity scores $\mathbf{\rho}_c\in\mathbb{R}^{B\times C\times M}$ and $\mathbf{\rho}_r\in\mathbb{R}^{B\times C\times N}$ between $\mathbf{X}$ and each prototype in $\mathcal{B}_c$ and $\mathcal{B}_r$ in the temporally latent space:
\begin{equation}
    \mathbf{\rho}_c = s(\phi{(\mathbf{X})}, \mathcal{B}_c),\quad \mathbf{\rho}_r = s(\phi{(\mathbf{X})}, \mathcal{B}_r),
\end{equation}  
where $s(\cdot, \cdot)$ denotes Pearson correlation as scale-invariant matching for temporal shape similarity. We then retrieve the most relevant prototypes via a dual-path selection strategy:
\begin{equation}
    \mathcal{I}_c = \arg \text{top-K}(\mathbf{\rho}_c),
\end{equation}
\begin{equation}
    \mathcal{I}_r = 
    \begin{cases}
    \arg \max (\rho_r), & \text{if } \max(\rho_r) > \epsilon \\
    \varnothing, & \text{otherwise}
    \end{cases}
\end{equation}
For common bank $\mathcal{B}_c$, we select $\text{top-K}$ prototypes to cover diverse recurring patterns. For rare bank $\mathcal{B}_r$, to enforce sparsity and precise memory, we select at most one prototype only when its maximum similarity exceeds a threshold $\epsilon = 0.3$. This prevents multiple weak rare matches from introducing noisy corrections. Only one serves as a sparsity prior rather than assuming that rare events are intrinsically unimodal.

Then we compute adaptive weights for the selected common and rare prototypes as:

\begin{equation}
    \omega_c = \text{Softmax}(\rho_c[\mathcal{I}_c] / \tau),\quad \omega_r = \rho_r[\mathcal{I}_r] / \tau,
\end{equation}
The retrieved pattern contributions are then weighted sum:
\begin{equation}
    \mathbf{z}_c = \sum_{k \in \mathcal{I}_c}\omega_c^{k} \cdot \mathcal{B}_c[\mathcal{I}_c], \quad \mathbf{z}_r = \omega_r \cdot \mathcal{B}_r[\mathcal{I}_r],
\end{equation}
Finally, we concatenate context-aware contributions with $\mathbf{h}$ and use a MLP layer to produce the enhanced prediction:
\begin{equation}
    \mathbf{\hat{Y}} = \mathbf{W}_o [\mathbf{h}; \mathbf{z}_c; \mathbf{z}_r],
\end{equation}
This adaptive fusion mechanism enables the model to selectively leverage context-specific common and rare patterns.

\subsection{Disentanglement-Guided Loss}

To prevent the two prototype banks from converging to homogeneous representations, we propose a Disentanglement-Guided Loss (DGLoss). It jointly optimizes the forecasting backbone and DDP by enforcing three key properties: pattern separation, rarity preservation, and common set diversity.

We first introduce the Separation Loss $\mathcal{L}_{\text{sep}}$ to encourage a clear functional divide between the common and rare banks:
\begin{equation}
    \mathcal{L_{\text{sep}}}=-\mathbb{E}\left[(1-\eta)\log\sigma(\triangle{\mathbf{\rho}})+\eta\log\sigma(-\triangle{\mathbf{\rho}})\right],
\end{equation}
where $\triangle{\mathbf{\rho}}=\max_i\mathbf{\rho}_c^i-\max_j\mathbf{\rho}_r^j$ is computed for each sample–variate pair as the difference between its maximum common-bank and rare-bank responses. $\eta = [(1-\max_i\mathbf{\rho}_c^i)/2]$ is soft rarity target, which reversely maps common bank coverage from the Pearson correlation range to [0, 1]. The expectation is taken over all sample–variate pairs. A well-covered input receives a low rarity target, whereas an input poorly explained by the common bank receives a high rarity target. Thus, $\mathcal{L}_{\text{sep}}$ discourages rare bank activation for frequent patterns and over-reliance on the common bank for rare patterns.


Then we propose a Rarity Preservation Loss $\mathcal{L}_{\text{rare}}$. With the contrastive restriction, $\mathcal{L}_{\text{rare}}$ prevents the rare prototypes from being contaminated or forgotten due to common patterns:
\begin{equation}
\mathcal{L}_{\mathrm{rare}}
=
-\frac{1}{|\mathcal{A}|}
\sum_{a\in\mathcal{A}}
\log
\frac{
\exp\left(\rho^r_{a,k_a}/\tau\right)
}{
\sum_{j=1}^{N}\exp\left(\rho^r_{a,j}/\tau\right)
},
\end{equation}
where $\mathcal{A}$ is the set of sample–variate positions that activate the rare path. $k_a$ is the rare prototype selected at position $a$. $\rho$ is defined in DPC. This input–prototype contrast directly aligns each activated rare prototype with its context while preserving rare-bank distinctiveness. If no sample in a mini-batch activates the rare bank, this term is set to zero.

In addition, we adopt a diversity loss $\mathcal{L}_{\text{div}}$ that leverages orthogonal constraints to promote diversity and reduce redundancy among common prototypes:
\begin{equation}
    \mathcal{L}_{\text{div}} = \frac{1}{M(M-1)} \sum_{i=1}^{M} \sum_{j \neq i}^{M} \left( \frac{ {\mathbf{p}_c^{i}}^\top \mathbf{p}_c^{j} }{ \|\mathbf{p}_c^{i}\| \|\mathbf{p}_c^{j}\| } \right)^2,
\end{equation}
where $M$ is the number of common prototypes and $\mathbf{p}_c^i$ is the $i$-th common prototype vector. 

Finally, we integrate these three loss terms as the Disentanglement-Guided Loss $\mathcal{L}_{\text{DGL}}$:
\begin{equation}
    \mathcal{L}_{\text{DGL}} = \lambda_\text{sep} \mathcal{L}_{\text{sep}} + \lambda_\text{rare} \mathcal{L}_{\text{rare}} + \lambda_\text{div} \mathcal{L}_{\text{div}},
\end{equation}
where $\lambda_\text{sep}, \lambda_\text{rare}, \lambda_\text{div}$ are balancing coefficients. The overall training objective is formulated as follows:
\begin{equation}
    \mathcal{L} = \mathcal{L}_\text{MSE} + \mathcal{L}_\text{DGL}
\end{equation}
By explicitly coordinating the separation, preservation, and diversity, DGLoss shapes a well-structured and role-specialized prototype memory for dynamic context-aware disentanglement.

\begin{table*}[!t]
\centering
\small
\setlength{\tabcolsep}{4pt}
\renewcommand{\arraystretch}{1.08}
\resizebox{\textwidth}{!}{
\begin{tabular}{@{}l cccc@{\quad} cccc@{\quad} cccc@{\quad} cccc@{\quad} cccc@{}}
\toprule
\multirow{2}{*}{Dataset}
& \multicolumn{4}{c}{iTransformer}
& \multicolumn{4}{c}{DLinear}
& \multicolumn{4}{c}{TimesNet}
& \multicolumn{4}{c}{TimeXer}
& \multicolumn{4}{c}{TimeBridge} \\
\cmidrule(lr){2-5} \cmidrule(lr){6-9} \cmidrule(lr){10-13} \cmidrule(lr){14-17} \cmidrule(l){18-21}
& \multicolumn{2}{c}{Ori} & \multicolumn{2}{c}{+DPAD}
& \multicolumn{2}{c}{Ori} & \multicolumn{2}{c}{+DPAD}
& \multicolumn{2}{c}{Ori} & \multicolumn{2}{c}{+DPAD}
& \multicolumn{2}{c}{Ori} & \multicolumn{2}{c}{+DPAD}
& \multicolumn{2}{c}{Ori} & \multicolumn{2}{c}{+DPAD} \\
Metric
& MSE & MAE & MSE & MAE
& MSE & MAE & MSE & MAE
& MSE & MAE & MSE & MAE
& MSE & MAE & MSE & MAE
& MSE & MAE & MSE & MAE \\
\midrule
ETTh1 & 0.452 & 0.446 & \textbf{0.446} & \textbf{0.439} & 0.461 & 0.457 & \textbf{0.449} & \textbf{0.450} & 0.479 & 0.466 & \textbf{0.476} & \textbf{0.463} & 0.460 & 0.452 & \textbf{0.442} & \textbf{0.441} & 0.450 & 0.443 & \textbf{0.432} & \textbf{0.431} \\
ETTh2 & 0.383 & 0.406 & \textbf{0.376} & \textbf{0.399} & 0.563 & 0.518 & \textbf{0.529} & \textbf{0.500} & 0.411 & 0.420 & \textbf{0.396} & \textbf{0.414} & 0.373 & 0.403 & \textbf{0.369} & \textbf{0.396} & 0.379 & 0.403 & \textbf{0.367} & \textbf{0.391} \\
ETTm1 & 0.407 & 0.412 & \textbf{0.401} & \textbf{0.405} & 0.404 & 0.407 & \textbf{0.394} & \textbf{0.404} & 0.418 & 0.417 & \textbf{0.400} & \textbf{0.409} & 0.390 & \textbf{0.395} & \textbf{0.386} & 0.394 & 0.387 & 0.399 & \textbf{0.379} & \textbf{0.387} \\
ETTm2 & 0.292 & 0.335 & \textbf{0.285} & \textbf{0.329} & 0.354 & 0.402 & \textbf{0.334} & \textbf{0.391} & 0.291 & 0.330 & \textbf{0.281} & \textbf{0.325} & \textbf{0.277} & 0.322 & 0.299 & \textbf{0.318} & \textbf{0.281} & 0.326 & 0.282 & \textbf{0.321} \\
\addlinespace[2pt]
Electricity & 0.175 & 0.264 & \textbf{0.170} & \textbf{0.261} & 0.225 & 0.319 & \textbf{0.214} & \textbf{0.305} & \textbf{0.193} & \textbf{0.293} & 0.196 & 0.296 & 0.201 & 0.275 & \textbf{0.170} & \textbf{0.268} & 0.171 & 0.266 & \textbf{0.165} & \textbf{0.256} \\
Exchange & 0.361 & 0.406 & \textbf{0.354} & \textbf{0.401} & \textbf{0.338} & \textbf{0.413} & 0.345 & 0.417 & 0.430 & 0.446 & \textbf{0.384} & \textbf{0.422} & 0.398 & 0.422 & \textbf{0.375} & \textbf{0.409} & 0.417 & 0.440 & \textbf{0.361} & \textbf{0.402} \\
Solar & 0.237 & 0.264 & \textbf{0.233} & \textbf{0.256} & 0.330 & 0.401 & \textbf{0.264} & \textbf{0.308} & 0.267 & 0.286 & \textbf{0.244} & \textbf{0.276} & 0.226 & 0.269 & \textbf{0.220} & \textbf{0.262} & 0.236 & 0.270 & \textbf{0.231} & \textbf{0.241} \\
Traffic & 0.422 & 0.282 & \textbf{0.416} & \textbf{0.278} & 0.688 & 0.427 & \textbf{0.613} & \textbf{0.385} & 0.633 & 0.333 & \textbf{0.511} & \textbf{0.318} & 0.466 & 0.286 & \textbf{0.463} & \textbf{0.284} & 0.912 & 0.544 & \textbf{0.444} & \textbf{0.286} \\
Weather & 0.260 & 0.281 & \textbf{0.256} & \textbf{0.276} & 0.265 & \textbf{0.316} & \textbf{0.259} & 0.317 & 0.256 & 0.283 & \textbf{0.244} & \textbf{0.272} & 0.242 & 0.272 & \textbf{0.239} & \textbf{0.270} & \textbf{0.258} & 0.281 & 0.256 & \textbf{0.273} \\
\midrule
PEMS03 & 0.260 & 0.327 & \textbf{0.188} & \textbf{0.283} & 0.279 & 0.377 & \textbf{0.197} & \textbf{0.301} & 0.153 & 0.254 & \textbf{0.125} & \textbf{0.229} & 0.180 & 0.279 & \textbf{0.134} & \textbf{0.248} & \textbf{0.153} & 0.262 & 0.155 & \textbf{0.259} \\
PEMS04 & 0.268 & 0.369 & \textbf{0.199} & \textbf{0.312} & 0.295 & 0.388 & \textbf{0.205} & \textbf{0.313} & 0.138 & 0.250 & \textbf{0.111} & \textbf{0.222} & 0.326 & 0.418 & \textbf{0.138} & \textbf{0.261} & 0.193 & 0.295 & \textbf{0.192} & \textbf{0.289} \\
PEMS07 & 0.109 & 0.214 & \textbf{0.104} & \textbf{0.207} & 0.328 & 0.394 & \textbf{0.223} & \textbf{0.310} & 0.109 & 0.216 & \textbf{0.105} & \textbf{0.206} & \textbf{0.088} & 0.192 & 0.089 & 0.192 & \textbf{0.128} & 0.232 & 0.130 & \textbf{0.230} \\
PEMS08 & 0.199 & 0.278 & \textbf{0.173} & \textbf{0.254} & 0.403 & 0.444 & \textbf{0.272} & \textbf{0.323} & 0.198 & 0.236 & \textbf{0.160} & \textbf{0.235} & 0.205 & \textbf{0.248} & \textbf{0.179} & 0.262 & 0.168 & 0.255 & \textbf{0.164} & \textbf{0.251} \\
\bottomrule
\end{tabular}}
\caption{Forecasting results for long-term and short-term tasks. Results are averaged over four prediction lengths: $\{12,24,48,96\}$ for PEMS and $\{96,192,336,720\}$ for the other datasets. “Ori” denotes the original backbone. The look-back length is fixed to 96. Better results between each backbone and its DPAD-enhanced version are highlighted in \textbf{bold}.}
\label{forecasting_results}
\end{table*}

\begin{table*}[!t]
\small
  \centering
  \setlength{\tabcolsep}{3.5pt}
\renewcommand{\arraystretch}{1.08}
  \resizebox{0.7\textwidth}{!}{
    \begin{tabular}{c|c|cccc|cccc|cccc}
    \toprule
    \multicolumn{2}{c|}{Dataset} & \multicolumn{4}{c|}{Electricity}    & \multicolumn{4}{c|}{ETTh2}    & \multicolumn{4}{c}{Weather} \\
    \midrule
    \multicolumn{2}{c|}{Forecast length} & 96    & 192   & 336   & 720   & 96    & 192   & 336   & 720   & 96    & 192   & 336   & 720 \\
    \midrule
    Backbone & MSE   & \underline{0.148} & 0.167 & 0.179 & 0.208 & 0.301 & 0.380 & 0.423 & 0.431 & 0.176 & 0.225 & 0.281 & 0.361 \\
    (\citeyear{iTransformer2023}) & MAE   & \underline{0.241} & \textbf{0.248} & \underline{0.271} & 0.298 & 0.350 & 0.399 & 0.431 & 0.447 & 0.216 & 0.257 & 0.299 & 0.353 \\
    \midrule
    RAFT & MSE   & 0.168 & 0.172 & 0.182 & 0.212 & \textbf{0.288} & 0.384 & 0.432 & 0.436 & 0.190 & 0.241 & 0.294 & 0.370 \\
    (\citeyear{han2025}) & MAE   & 0.269 & 0.270 & 0.281 & 0.304 & \textbf{0.341} & 0.402 & 0.443 & 0.460 & 0.237 & 0.281 & 0.323 & 0.374 \\
    \midrule
    FreDF & MSE   & 0.158 & 0.166 & 0.188 & 0.225 & 0.290 & 0.376 & \underline{0.418} & 0.426 & \underline{0.171} & \underline{0.224} & \underline{0.282} & \underline{0.359} \\
    (\citeyear{FreDF2025}) & MAE   & 0.248 & 0.259 & 0.277 & 0.308 & \underline{0.342} & \underline{0.394} & \underline{0.429} & \underline{0.443} & \underline{0.209} & \textbf{0.255} & \underline{0.298} & \underline{0.350} \\
    \midrule
    SARAF & MSE   & 0.153 & \textbf{0.160} & \underline{0.179} & \textbf{0.195} & \underline{0.288} & \textbf{0.372} & 0.431 & \underline{0.423} & 0.180 & 0.230 & 0.285 & 0.366 \\
    (\citeyear{saraf2026}) & MAE   & 0.257 & 0.260 & 0.278 & \underline{0.290} & 0.342 & 0.394 & 0.441 & 0.451 & 0.235 & 0.279 & 0.323 & 0.372 \\
    \midrule
    UEC-STD & MSE   & 0.199 & 0.212 & 0.227 & 0.267 & 0.294 & 0.385 & 0.423 & 0.434 & 0.183 & 0.230 & 0.285 & 0.361 \\
    (\citeyear{uec2026}) & MAE   & 0.292 & 0.304 & 0.318 & 0.348 & 0.346 & 0.399 & 0.433 & 0.452 & 0.223 & 0.262 & 0.301 & 0.351 \\
    \midrule
    {\textbf{DPAD}} & MSE   & \textbf{0.147} & \underline{0.162} & \textbf{0.172} & \underline{0.201} & 0.293 & \underline{0.374} & \textbf{0.416} & \textbf{0.422} & \textbf{0.170} & \textbf{0.223} & \textbf{0.279} & \textbf{0.355} \\
    {(\textbf{Ours})}& MAE   & \textbf{0.234} & \underline{0.251} & \textbf{0.271} & \textbf{0.289} & 0.345 & \textbf{0.393} & \textbf{0.425} & \textbf{0.435} & \textbf{0.208} & \underline{0.256} & \textbf{0.295} & \textbf{0.347} \\
    \bottomrule
    \end{tabular}
    }

\caption{Comparison with representative enhancement strategies with same configurations on Electricity, ETTh2, and Weather datasets, we use iTransformer as the backbone. The best results are highlighted in \textbf{bold}, and the second-best results are underlined.}
\label{comp_es}
\end{table*}

\section{Experiments}
To comprehensively verify the performance and effectiveness of the proposed DPAD, we conduct extensive experiments on both long-term and short-term forecasting tasks.

\subsection{Experimental Setup}
\subsubsection{Datasets.} We conduct long-term forecasting experiments on diverse real-world datasets, including ETT (ETTh1, ETTh2, ETTm1, ETTm2), Electricity, Exchange, Solar-Energy, Weather and Traffic. Meanwhile, we conduct short-term forecasting experiments on PEMS (PEMS03, PEMS04, PEMS07, PEMS08) dataset.

\subsubsection{Baselines.} We evaluate DPAD on five state-of-the-art forecasting backbones with diverse architectures, including transformer-based models: iTransformer \cite{iTransformer2023}, TimeXer \cite{TimeXer2024}, TimeBridge \cite{liu2025t}; MLP-based models: DLinear \cite{Ailing2023}; and CNN-based models: TimesNet \cite{TimesNet2023}.

\subsection{Main Results}
We present main results for long-term and short-term forecasting on thirteen real-world datasets for five SOTA forecasting models in Table \ref{forecasting_results}, where the lower MSE and MAE indicate better forecasting performance. DPAD demonstrates consistent improvements across all backbone models, achieving the lowest MSE and MAE in most cases. Notably, DPAD achieves an average MSE reduction of 12.6\% for DLinear and 9.3\% for iTransformer, and follows a similar trend for other backbones. The improvements are especially clear on complex datasets such as Traffic and PEMS. These consistent gains across heterogeneous backbones suggest that role-specialized prototype memory provides complementary pattern information beyond backbone-specific representations.

\subsection{Comparison with other Enhancement Strategies}
To further validate the effectiveness of DPAD, we compare it against several representative enhancement strategies. RAFT \cite{han2025} employs retrieval-augmented generation to enhance forecasting. FreDF \cite{FreDF2025} forecasts in the frequency domain to avoid label autocorrelation modeling. SARAF \cite{saraf2026} leverages stationarity-aware retrieval for forecasting. UEC-STD \cite{uec2026} corrects AR time series forecasts via seasonal-trend decomposition. As shown in Table \ref{comp_es}, DPAD achieves superior forecasting performance in most cases. This advantage mainly comes from its unified framework for context-aware pattern disentanglement and role-specialized memory, which addresses the limitations of role-agnostic enhancement more directly than other strategies. Thus DPAD demonstrates stronger generalization across diverse forecasting scenarios.

\subsection{Ablation Studies}
To validate the contribution of each component in DPAD, we conduct systematic ablation studies on Electricity, Weather, Traffic and Solar. The results are reported in Tables~\ref{ablation1}--\ref{ablation3}.

For the Dynamic Dual-Prototype bank (DDP), we evaluate: \ding{172} without the whole DDP, \ding{173} with only the Common Bank, \ding{174} with only the Rare Bank. Table \ref{ablation1} shows removing DDP generally degrades performance. Interestingly, using either bank alone provides limited or even negative gains, whereas combining both usually achieves the best or competitive results. This indicates that a single-bank design is insufficient to provide meaningful role contrast, while the dual-bank structure allows the separation loss to encourage specialization between common-like and rare-like activations.
\begin{table}[htbp]
  \centering
  \resizebox{\columnwidth}{!}{
    \begin{tabular}{c|c|c|cc|cc|cc|cc}
    \toprule
    \multirow{2}[0]{*}{Case} & \multirow{2}[0]{*}{\makecell{Common\\Bank}} & \multirow{2}[0]{*}{\makecell{Rare\\Bank}} & \multicolumn{2}{c|}{\multirow{1}{*}{Electricity}} & \multicolumn{2}{c|}{\multirow{1}{*}{Weather}} & \multicolumn{2}{c|}{\multirow{1}{*}{Traffic}} & \multicolumn{2}{c}{\multirow{1}{*}{Solar}} \\
    \cmidrule(lr){4-5} \cmidrule(lr){6-7} \cmidrule(lr){8-9} \cmidrule(lr){10-11}
          &  &   & MSE   & MAE   & MSE   & MAE   & MSE   & MAE   & MSE   & MAE \\
    \midrule
    \ding{172}     & $\times$ & $\times$  & 0.175  & 0.264  & 0.260  & 0.281  & 0.422  & 0.282  & 0.237  & 0.262  \\
    \midrule
    \ding{173}     & $\checkmark$  & $\times$  & 0.178  & 0.269  & 0.262  & 0.282  & 0.462  & 0.314  & 0.240  & 0.266  \\
    \midrule
    \ding{174}     & $\times$ & $\checkmark$  & 0.183  & 0.272  & 0.263  & 0.282  & 0.423  & 0.283  & 0.237  & 0.262  \\
    \midrule
    \textbf{\ding{175} (Ours)}     & \textbf{$\checkmark$} & \textbf{$\checkmark$} & \textbf{0.170}  & \textbf{0.261}  & \textbf{0.256}  & \textbf{0.276}  & \textbf{0.416}  & \textbf{0.278}  & \textbf{0.233}  & \textbf{0.256}  \\
    \bottomrule
    \end{tabular}
    }
    \caption{Ablation study on DDP to validate the efficacy of Common Pattern Bank and Rare Pattern Bank. $\checkmark$ and $\times$ indicate with and without certain components respectively. All the results are averaged from 4 different prediction lengths.}
    \label{ablation1}
\end{table}

For Dual-Path Context-aware routing (DPC), we compare: \ding{172} element-wise sum fusion, replacing adaptive retrieval with a simple summation of prototype vectors, and \ding{173} fixed average fusion, substituting it with static average weighting. Table \ref{ablation2} shows that both variants underperform full DPC. This indicates that static fusion lacks dynamic selection and weighting mechanisms based on input context. In contrast, DPC dynamically retrieves and weights role-specific prototypes, enabling the model to selectively leverage relevant patterns.

\begin{table}[htbp]
  \centering
  \resizebox{\columnwidth}{!}{
    \begin{tabular}{c|cc|cc|cc|cc}
    \toprule
    \multirow{2}[0]{*}{Case} & \multicolumn{2}{c|}{\multirow{1}{*}{Electricity}} & \multicolumn{2}{c|}{\multirow{1}{*}{Weather}} & \multicolumn{2}{c|}{\multirow{1}{*}{Traffic}} & \multicolumn{2}{c}{\multirow{1}{*}{Solar}} \\
    \cmidrule(lr){2-3} \cmidrule(lr){4-5} \cmidrule(lr){6-7} \cmidrule(lr){8-9}
       & MSE   & MAE   & MSE   & MAE   & MSE   & MAE   & MSE   & MAE \\
    \midrule
    \ding{172} w/ Additive   & 0.179  & 0.268  & 0.262  & 0.282  & 0.422  & 0.282  & 0.237  & 0.268  \\
    \midrule
    \ding{173} w/ Mean    & 0.177  & 0.269  & 0.260  & 0.281  & 0.439  & 0.298  & 0.237  & 0.262  \\
    \midrule
    \textbf{\ding{174} Full DPC (Ours)}   & \textbf{0.170}  & \textbf{0.261}  & \textbf{0.256}  & \textbf{0.276}  & \textbf{0.416}  & \textbf{0.278}  & \textbf{0.233}  & \textbf{0.256}  \\
    \bottomrule
    \end{tabular}
    }
    \caption{Ablation study on DPC to compare adaptive routing with different static fusion strategies.}
    \label{ablation2}
\end{table}

For Disentanglement-Guided Loss (DGLoss), we conduct: \ding{172} without the whole DGLoss, \ding{173} without separation loss $\mathcal{L}_{\text{sep}}$, \ding{174} without rarity preservation loss $\mathcal{L}_{\text{rare}}$, and \ding{175} without diversity loss $\mathcal{L}_{\text{div}}$. Table \ref{ablation3} shows that removing any individual loss term or the entire DGLoss leads to a performance drop. Without explicit supervision for separation, diversity, and rarity preservation, our DDP may lose its designated specialization or provide insufficient pattern coverage. These results confirm that all three objectives are necessary for learning a well-organized dual-prototype memory.

In summary, the ablation studies validate that all three components of DPAD are essential and complementary. DDP organizes temporal patterns into role-specialized prototype banks, DPC selectively leverages the stored patterns according to the input context, and DGLoss maintains role specialization and comprehensive prototype coverage. Removing any component weakens this coordinated process, demonstrating that DPAD's effectiveness stems from the synergistic interaction of these three designs.

\begin{table}[htbp]
  \centering
  \resizebox{\columnwidth}{!}{
    \begin{tabular}{c|c|c|c|cc|cc|cc|cc}
    \toprule
    \multirow{2}[0]{*}{Case} & \multirow{2}[0]{*}{$\mathcal{L}_{\text{sep}}$} & \multirow{2}[0]{*}{$\mathcal{L}_{\text{rare}}$} &  \multirow{2}[0]{*}{$\mathcal{L}_{\text{div}}$} & \multicolumn{2}{c|}{\multirow{1}{*}{Electricity}} & \multicolumn{2}{c|}{\multirow{1}{*}{Weather}} & \multicolumn{2}{c|}{\multirow{1}{*}{Traffic}} & \multicolumn{2}{c}{\multirow{1}{*}{Solar}} \\
    \cmidrule(lr){5-6} \cmidrule(lr){7-8} \cmidrule(lr){9-10} \cmidrule(lr){11-12} 
          &  & &  & MSE   & MAE   & MSE   & MAE   & MSE   & MAE   & MSE   & MAE \\
    \midrule
    \ding{172}     & $\times$ & $\times$ & $\times$  & 0.180  & 0.271  & 0.264  & 0.283  & 0.441  & 0.298  & 0.241  & 0.267  \\
    \midrule
    \ding{173}     & $\times$  & $\checkmark$ & $\checkmark$  & 0.178  & 0.269  & 0.262  & 0.281  & 0.464  & 0.322  & 0.239  & 0.266  \\
    \midrule
    \ding{174}     & $\checkmark$ & $\times$ & $\checkmark$  & 0.179  & 0.268  & 0.261  & 0.282  & 0.450  & 0.308  & 0.239  & 0.266  \\
    \midrule
    \ding{175}     & $\checkmark$ & $\checkmark$ & $\times$  & 0.177  & 0.269  & 0.262  & 0.283  & 0.441  & 0.299  & 0.238  & 0.265  \\
    \midrule
    \textbf{\ding{176} (Ours)}     & \textbf{$\checkmark$} & \textbf{$\checkmark$} & \textbf{$\checkmark$} & \textbf{0.170}  & \textbf{0.261}  & \textbf{0.256}  & \textbf{0.276}  & \textbf{0.416}  & \textbf{0.278}  & \textbf{0.233}  & \textbf{0.256}  \\
    \bottomrule
    \end{tabular}
    }
  \caption{Ablation study on DGLoss to assess the contribution of each loss component.}
  \label{ablation3}
\end{table}

\subsection{Model Analysis}

\begin{figure*}[!t]
  \centering
    \begin{subfigure}[t]{0.31\textwidth}
        \includegraphics[width=\textwidth]{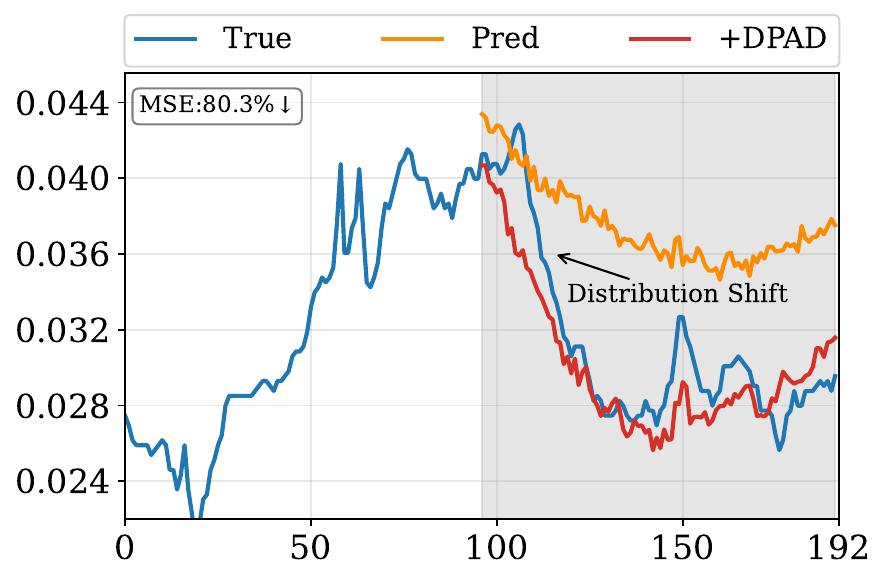}
        \caption{Distribution Shifts}
    \end{subfigure}
    \hspace{-0.05cm}
    \begin{subfigure}[t]{0.31\textwidth}
        \includegraphics[width=\textwidth]{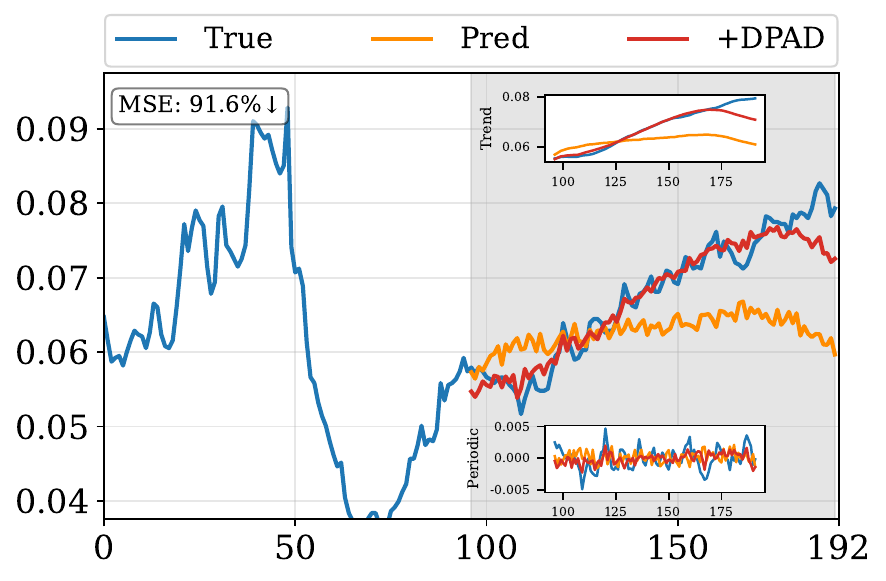}
        \caption{Intertwined Patterns}
    \end{subfigure}
    \hspace{-0.05cm}
    \begin{subfigure}[t]{0.31\textwidth}
        \includegraphics[width=\textwidth]{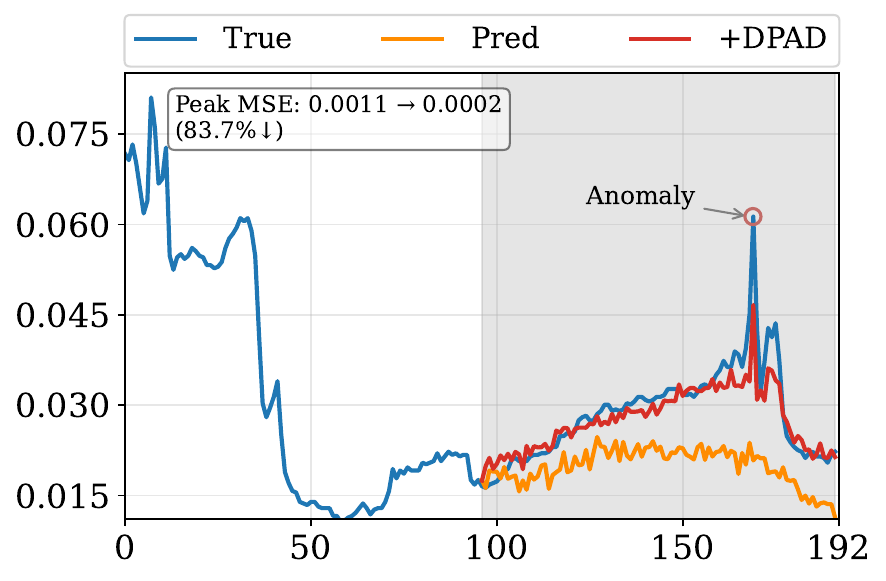}
        \caption{Rare Events}
    \end{subfigure}
    \caption{Case studies on failure scenarios:a) distribution shifts; b) intertwined patterns; c) rare events.}
    \label{case_studies}
\end{figure*}
\begin{figure*}[!t]
  \centering
    \includegraphics[width=\textwidth]{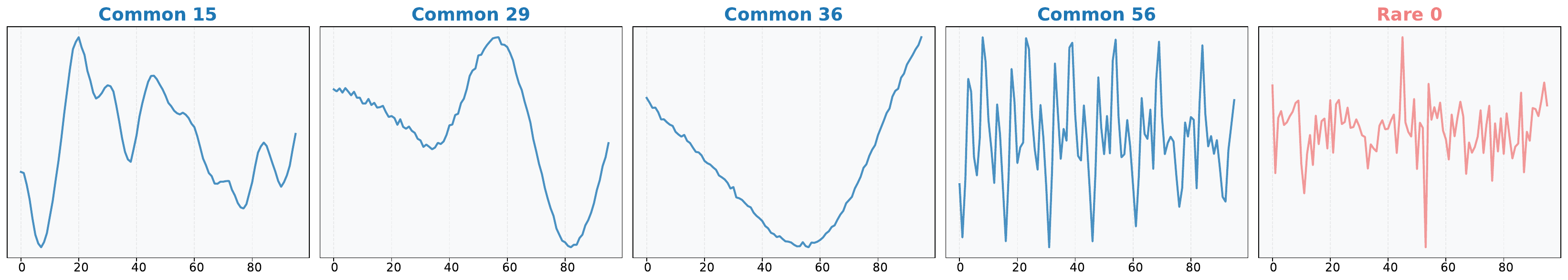}
    \caption{Visualization of learned prototypes in DDP on the Electricity dataset.}
    \label{proto_visual}
\end{figure*}

\subsubsection{Case Studies On Failure Scenarios}

To examine whether DPAD can dynamically distinguish and utilize common and rare patterns under different contexts, we select representative sequences from the Weather dataset involving distribution shifts, intertwined patterns, and rare events. As shown in Figure \ref{case_studies}, for (a) distribution shifts, the backbone fails to adapt after the temporal shift, while DPAD retrieves relevant common structures as stable references and adjusts the prediction accordingly. For (b) intertwined patterns, a linear trend coexists with multi-frequency periodic components, the backbone inadequately captures the mixed structure, whereas DPAD retrieves multiple relevant prototypes through context-aware routing, improving the modeling of these intertwined common structures. For (c) rare events, characterized by a sudden temperature spike, the backbone yields a large peak error, while DPAD selectively activates rare-pattern information and reduces the MSE in the peak region by 83.7\%. Collectively, these case studies demonstrate that DPAD dynamically organizes and utilizes both common and rare patterns according to their forecasting roles and the current input context.



\subsubsection{Prototype Visualization.} 
To intuitively understand what DDP learns, we visualize the temporal forms of several representative prototypes from the common and rare banks on the Electricity dataset. As shown in Figure \ref{proto_visual}, prototypes in both banks evolve dynamically during training while exhibiting distinct temporal characteristics. Common prototypes develop smooth and structured shapes, including combinations of trends and seasonalities, suggesting that they capture diverse prevalent dynamics. In contrast, rare prototypes exhibit abrupt fluctuations and sudden shifts, indicating their adaptation to infrequent temporal variations. This distinction supports the intended role specialization of the two banks and provides qualitative evidence that DDP organizes heterogeneous patterns according to their forecasting roles

\subsubsection{Increasing Look-Back Length.}
\begin{figure}[!t]
  \centering
  \includegraphics[width=\linewidth]{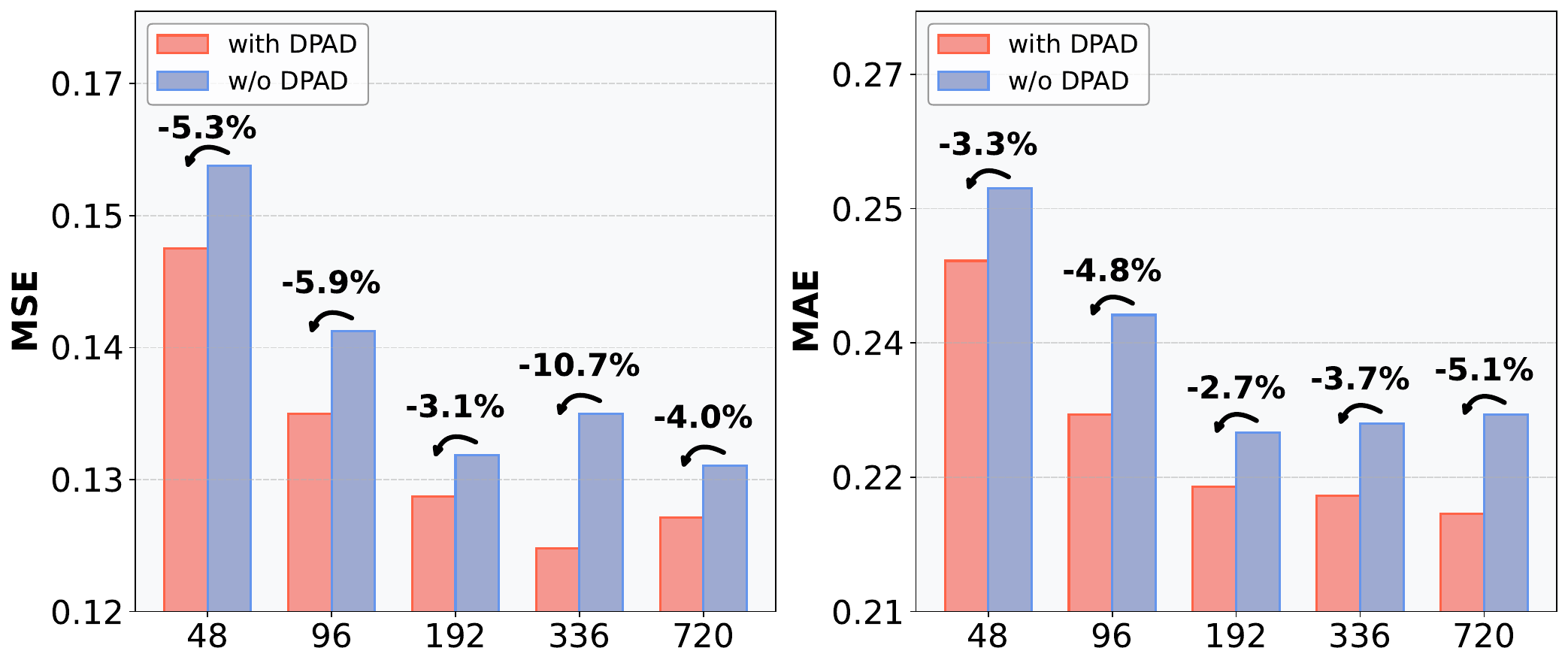}  
  \caption{Forecasting results (MSE and MAE) with varying look-back length $\{48, 96, 192, 336, 720\}$ on Electricity dataset. Prediction length is fixed to 96.}
  \label{input_length}
\end{figure}

A longer look-back sequence provides richer historical information in principle, but also introduces more heterogeneous and intertwined temporal patterns. To evaluate whether DPAD can disentangle and leverage long-range context, we conduct experiments under increasing look-back length $\{48, 96, 192, 336, 720\}$ on the Electricity dataset, using TimeBridge as backbone. As shown in Figure \ref{input_length}, DPAD maintains performance gains across different look-back lengths. Vanilla backbones may suffer from information overload or struggle to focus on useful long-term dependencies. By organizing intertwined patterns into role-specialized prototype memory and selectively retrieving relevant prototypes, DPAD helps the backbone utilize extended historical contexts more effectively. These results further demonstrate its ability to dynamically distinguish and leverage relevant common and rare patterns under varying context lengths.

\begin{table}[!t]
  \centering
\resizebox{\linewidth}{!}{
  \begin{tabular}{c|c|c|c|c|c|c}
    \toprule
    Metrics & \multicolumn{3}{c|}{Running Time} & \multicolumn{3}{c}{Memory Footprint}  \\
    \midrule
    Methods & \multicolumn{1}{c|}{ori} & \multicolumn{1}{c|}{+DPAD} & \multicolumn{1}{c|}{$\Delta\%$} & \multicolumn{1}{c|}{ori} & \multicolumn{1}{c}{+DPAD} & \multicolumn{1}{c}{$\Delta\%$}\\
    \midrule
    iTransformer & 17ms     & 20ms & 17.6\%     & 782MB     & 820MB & 4.8\% \\
    TimeXer & 119ms     & 126ms  & 5.8\%   & 4634MB     & 4746MB & 2.4\% \\
    TimeBridge & 98ms     & 107ms  & 9.1\%   & 4240MB     & 4334MB& 2.2\% \\
    DLinear & 7ms     & 11ms   & 57.1\%  & 62MB     & 90MB & 45.1\% \\
    TimesNet & 262ms     & 266ms   & 1.5\%  & 3332MB     & 3470MB  & 4.1\% \\
    \bottomrule
    \end{tabular}
    }
    \caption{Comparison of efficiency analysis for backbone models with and without DPAD. The results are on the Electricity dataset with batch size 16. The look-back length and prediction length are fixed to 96. Running time means running time per iteration. $\Delta\%$ denotes the relative change rate.}
  \label{efficiency}
\end{table}

\subsubsection{Efficiency Analysis.}
We conduct the efficiency analysis for the backbone with and without DPAD. As shown in Table \ref{efficiency}, DPAD introduces modest overhead for most backbones. The relative increases in running time and memory footprint are generally modest, which are below 10\% for most Transformer or CNN-based backbones. Notably, iTransformer incurs a 17.6\% increase in running time, while its memory overhead remains below 5\%. Although DLinear shows a higher relative increase due to its extremely small base cost, the absolute added overhead remains minimal. This indicates that DPAD is a practical and efficient enhancement module, delivering significant performance improvements while incurring only a small fraction of additional computational cost.

\section{Conclusion}
In this paper, we study the problem of role-agnostic pattern organization in time series forecasting. The key challenge is not merely to model more temporal patterns, but to organize and selectively leverage heterogeneous patterns according to their distinct forecasting roles. To address this challenge, we propose the Dual-Prototype Adaptive Disentanglement (DPAD) framework, a model-agnostic auxiliary method that enhances the backbone model through a Dynamic Dual-Prototype bank (DDP), Dual-Path Context-aware routing (DPC), and a Disentanglement-Guided Loss (DGLoss). Extensive experiments validate that DPAD consistently improves forecasting performance while introducing acceptable absolute overhead for most evaluated backbones. Future work may explore more interpretable routing strategies and extend role-specialized prototype memory to multimodal temporal representations.

\bibliography{aaai2027}


\appendix
\section{Implementation Details}

\subsection{Dataset Descriptions}
We conduct long-term forecasting experiments on 6 real-world datasets, including (1) ETT (ETTh1, ETTh2, ETTm1, ETTm2) contains 7 features of electricity transformer data from July 2016 to July 2018, which was sampled at hourly (ETTh1, ETTh2) and 15-minute (ETTm1, ETTm2) intervals. (2) Electricity (ECL) records hourly electricity consumption data of 321 clients from 2012 to 2014. (3) Exchange collects the daily exchange-rate data from eight different countries. (4) Solar-Energy contains the solar power production of 137 PV plants in 2006, which is sampled every 10 minutes. (5) Weather includes 21 meteorological factors collected every 10 minutes from the Max Planck Biogeochemistry Institute's Weather Station in 2020. (6) Traffic contains hourly road occupancy rates from 862 sensors on San Francisco freeways. Meanwhile, we conduct short-term forecasting experiments on PEMS (PEMS03, PEMS04, PEMS07, PEMS08), which contains the public traffic network data in California with a 5-minute interval. The details of datasets are listed in Table \ref{datasets}.

\begin{table*}[htbp]
	\centering
	\resizebox{0.9\textwidth}{!}{
	\begin{tabular}{c c c c c c c}
		\toprule
		\textbf{Name} & \textbf{Domain} & \textbf{Length} & \textbf{Num} & \textbf{Prediction Length} & \textbf{Dataset Size} & \textbf{Freq. (m)} \\
		\toprule
		ETTh1 & Temperature & 14400 & 7 & \{96,192,336,720\} & (8545,2881,2881) & 60 \\
		\midrule
		ETTh2 & Temperature & 14400 & 7 & \{96,192,336,720\} & (8545,2881,2881) & 60 \\
		\midrule
		ETTm1 & Temperature & 57600 & 7 & \{96,192,336,720\} & (34465,11521,11521) & 15 \\
		\midrule
		ETTm2 & Temperature & 57600 & 7 & \{96,192,336,720\} & (34465,11521,11521) & 15 \\
		\midrule
		Electricity & Electricity & 26304 & 321 & \{96,192,336,720\} & (18317,2633,5261) & 60 \\
		\midrule
		Exchange & Exchange Rate & 7588 & 8 & \{96,192,336,720\} & (5120,665,1422) & 1440 \\
		\midrule
		Traffic & Road Occupancy & 17544 & 862 & \{96,192,336,720\} & (12185,1757,3509) & 60 \\
		\midrule
		Weather & Weather & 52696 & 21 & \{96,192,336,720\} & (36792,5271,10540) & 10 \\
		\midrule
		Solar-Energy & Energy & 52179 & 137 & \{96,192,336,720\} & (36601,5161,10417) & 10 \\
		\midrule
		PEMS03 & Traffic Flow & 26208 & 358 & \{12,24,48,96\} & (15617,5135,5135) & 5 \\
		\midrule
		PEMS04 & Traffic Flow & 16992 & 307 & \{12,24,48,96\} & (10172,3375,3375) & 5 \\
		\midrule
		PEMS07 & Traffic Flow & 28224 & 883 & \{12,24,48,96\} & (16711,5622,5622) & 5 \\
		\midrule
		PEMS08 & Traffic Flow & 17856 & 170 & \{12,24,48,96\} & (10690,3548,3548) & 5 \\
		\bottomrule
	\end{tabular}}
        \caption{Dataset Descriptions. "Num" denotes the number of variables. "Dataset Size" is organized in (Train, Validation, Test). “Freq.” denotes the sampling interval in minutes.}
	\label{datasets}
\end{table*}

\subsection{Implementation Details.} All experiments are implemented in PyTorch, and conducted on a single NVIDIA RTX 4090 24 GB GPU. We use the ADAM optimizer with an initial learning rate $10^{-3}$ and optimize the forecasting objective using L2 loss. For fair evaluation, we use the framework of TimesNet, and all the baselines are implemented based on the configurations of original paper and official code.

\subsection{Metrics Details.}
To evaluate forecasting performance for TSF, we utilize the mean square error (MSE) and mean absolute error (MAE), where lower values indicate better performance. The calculations of metrics are as follows:
\begin{equation}
	\label{metrics MSE}
	\textbf{MSE} = \frac{1}{n}\sum_{i=1}^{n}(\mathbf{Y}_i - \hat{\mathbf{Y}}_i)^2, \quad \textbf{MAE} = \frac{1}{n}\sum_{i=1}^{n}|\mathbf{Y}_i - \hat{\mathbf{Y}}_i|.
\end{equation}

\section{Zero-Shot Forecasting}
To further evaluate the generalization and robustness of our DPAD on unseen datasets, we conduct zero-shot forecasting experiments. Following prior works, we sequentially use ETTh1, ETTh2, ETTm1, and ETTm2 as source datasets, and the remaining datasets as target datasets.

\begin{table*}[!t]
  \centering
  \resizebox{0.95\textwidth}{!}{
    \begin{tabular}{c|c|cc|cc|cc|cc|cc|cc|cc|cc}
    \toprule
    \multicolumn{2}{c|}{Methods} & \multicolumn{2}{c}{iTransformer} & \multicolumn{2}{c|}{+DPAD} & \multicolumn{2}{c}{DLinear} & \multicolumn{2}{c|}{+DPAD}  & \multicolumn{2}{c}{TimesNet} & \multicolumn{2}{c|}{+DPAD} & \multicolumn{2}{c}{TimeXer} & \multicolumn{2}{c}{+DPAD}\\
    \midrule
    Source & Target & MSE   & MAE   & MSE   & MAE   & MSE   & MAE   & MSE   & MAE   & MSE   & MAE   & MSE   & MAE   & MSE   & MAE   & MSE   & MAE \\
    \midrule
    \multirow{3}[2]{*}{ETTh1} & ETTh2 & 0.419  & 0.428  & \textbf{0.416 } & \textbf{0.425 } & 0.495  & 0.487  & \textbf{0.489 } & \textbf{0.481 } & 0.480  & 0.459  & \textbf{0.439 } & \textbf{0.443 } & 0.426  & 0.429  & \textbf{0.420 } & \textbf{0.425 } \\
          & ETTm1 & \textbf{0.822 } & \textbf{0.585 } & 0.834  & 0.596  & 0.749  & 0.579  & \textbf{0.746 } & \textbf{0.576 } & 0.849  & 0.575  & \textbf{0.754 } & \textbf{0.571 } & 0.832  & 0.591  & \textbf{0.798 } & \textbf{0.575 } \\
          & ETTm2 & 0.342  & 0.375  & \textbf{0.339 } & \textbf{0.373 } & 0.409  & 0.454  & \textbf{0.405 } & \textbf{0.451 } & 0.368  & 0.390  & \textbf{0.339 } & \textbf{0.370 } & 0.344  & 0.374  & \textbf{0.341 } & \textbf{0.370 } \\
    \midrule
    \multirow{3}[2]{*}{ETTh2} & ETTh1 & 0.694  & 0.581  & \textbf{0.632 } & \textbf{0.547 } & 0.544  & 0.508  & \textbf{0.539 } & \textbf{0.504 } & 0.802  & 0.623  & \textbf{0.769 } & \textbf{0.602 } & 0.695  & 0.577  & \textbf{0.666 } & \textbf{0.558 } \\
          & ETTm1 & 0.980  & 0.631  & \textbf{0.904 } & \textbf{0.606 } & \textbf{0.768 } & \textbf{0.594 } & 0.783  & 0.605  & 1.059  & 0.658  & \textbf{1.043 } & \textbf{0.644 } & 1.428  & 0.748  & \textbf{1.412 } & \textbf{0.734 } \\
          & ETTm2 & 0.358  & 0.389  & \textbf{0.346 } & \textbf{0.378 } & 0.537  & 0.541  & \textbf{0.526 } & \textbf{0.532 } & \textbf{0.369 } & \textbf{0.394 } & 0.380  & 0.404  & 0.387  & 0.408  & \textbf{0.377 } & \textbf{0.402 }  \\
    \midrule
    \multirow{3}[2]{*}{ETTm1} & ETTh1 & 0.718  & 0.571  & \textbf{0.670 } & \textbf{0.549 } & 0.614  & 0.530  & \textbf{0.602 } & \textbf{0.522 } & 0.952  & 0.694  & \textbf{0.930 } & \textbf{0.660 } & 0.960  & 0.662  & \textbf{0.948 } & \textbf{0.643 } \\
          & ETTh2 & 0.482  & 0.465  & \textbf{0.476 } & \textbf{0.459 } & 0.500  & 0.492  & \textbf{0.492 } & \textbf{0.479 } & 0.566  & 0.515  & \textbf{0.555 } & \textbf{0.507 } & 0.591  & 0.522  & \textbf{0.578 } & \textbf{0.507 } \\
          & ETTm2 & 0.326  & 0.362  & \textbf{0.312 } & \textbf{0.351 } & \textbf{0.318 } & \textbf{0.356 } & 0.337  & 0.367  & 0.357  & 0.380  & \textbf{0.340 } & \textbf{0.369 } & \textbf{0.336 } & \textbf{0.364 } & 0.341  & 0.371 \\
    \midrule
    \multirow{3}[2]{*}{ETTm2} & ETTh1 & 1.214  & 0.748  & \textbf{0.980 } & \textbf{0.676 } & 0.634  & 0.554  & \textbf{0.611 } & \textbf{0.523 } & 0.993  & 0.655  & \textbf{0.872 } & \textbf{0.619 } & 0.734  & 0.578  & \textbf{0.709 } & \textbf{0.576 } \\
          & ETTh2 & 0.553  & 0.507  & \textbf{0.519 } & \textbf{0.487 } & \textbf{0.508 } & \textbf{0.499 } & 0.519  & 0.506  & 0.528  & 0.487  & \textbf{0.490 } & \textbf{0.472 } & 0.457  & 0.453  & \textbf{0.436 } & \textbf{0.425 } \\
          & ETTm1 & 0.825  & 0.589  & \textbf{0.804 } & \textbf{0.579 } & 0.561  & 0.498  & \textbf{0.544 } & \textbf{0.479 } & 0.709  & 0.554  & \textbf{0.701 } & \textbf{0.548 } & 0.647  & 0.531  & \textbf{0.631 } & \textbf{0.525 } \\
    \bottomrule
    \end{tabular}
    }
    \caption{Zero-shot forecasting results on the ETT datasets. The forecasting length is set to 96. And the better results are highlighted in \textbf{bold}. }    
    \label{zero-shot}
\end{table*}
Table \ref{zero-shot} reports zero-shot results with four backbone models, which consistently demonstrate that models enhanced with our DPAD framework achieve superior performance in most cases. These improvements suggest that organizing temporal patterns according to their forecasting roles can improve the transferability of the learned prototype memory. In particular, the common bank captures recurring temporal structures that may be shared across related ETT datasets, while the dual-path routing mechanism selectively retrieves relevant common and rare patterns according to the target input context. Consequently, DPAD can provide context-specific enhancement on unseen datasets without retraining, demonstrating that role-aware pattern organization and context-aware utilization remain effective beyond the training distribution.


\section{Hyperparameter Sensitivity Analysis}
\subsection{Size of Prototype Banks}
\begin{figure*}[!t]
  \centering
    \begin{subfigure}{0.45\textwidth}
        \includegraphics[width=\columnwidth]{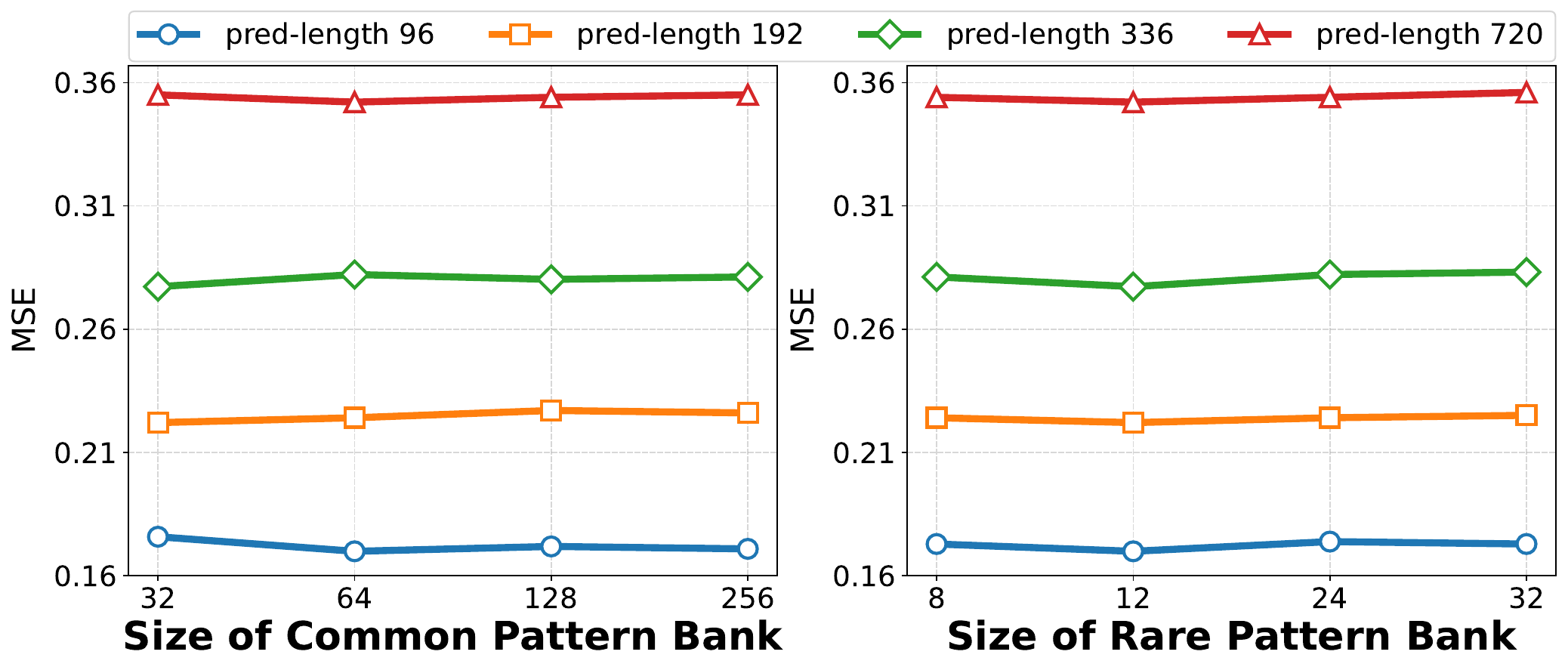}
        \caption{On Weather dataset.}
        \label{prototype1}
    \end{subfigure}
    \hspace{-0.1cm}
    \begin{subfigure}{0.45\textwidth}
        \includegraphics[width=\columnwidth]{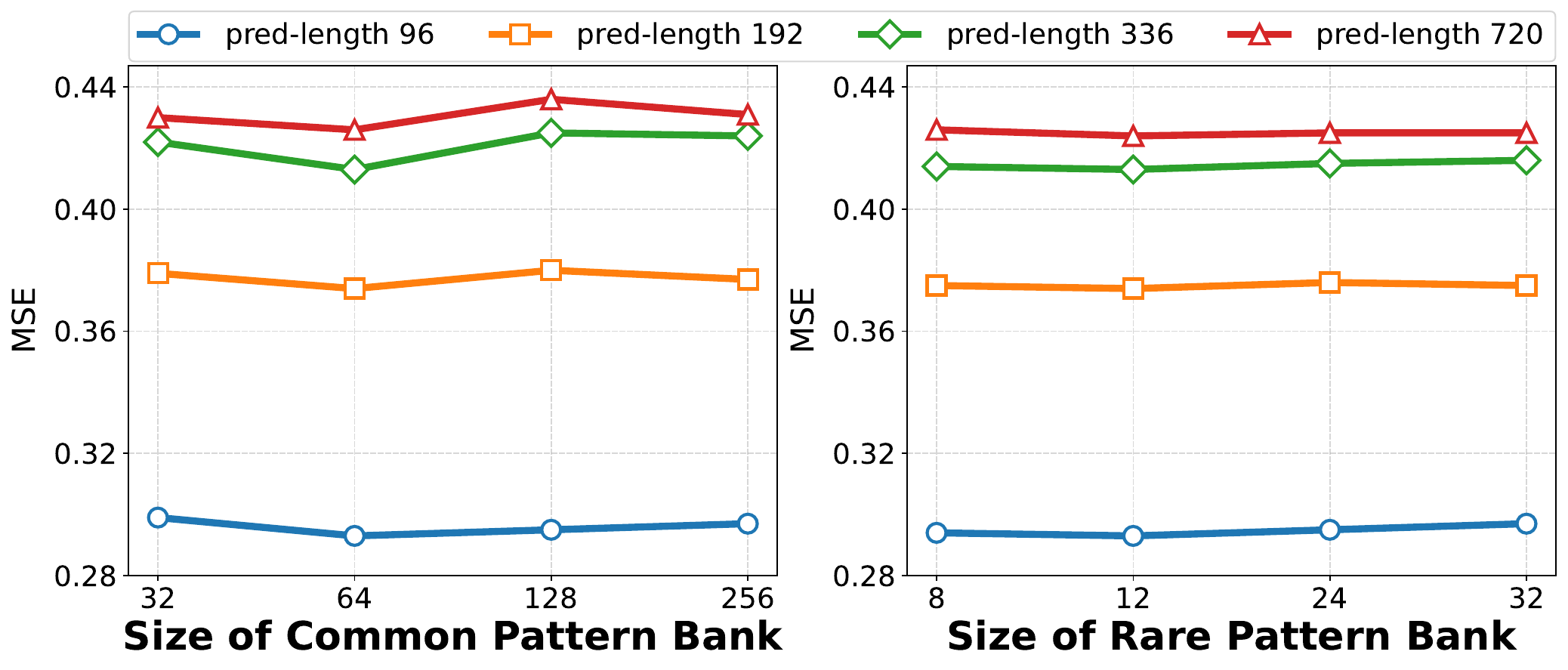}
        \caption{On ETTh2 dataset.}
        \label{prototype2}
    \end{subfigure}
    \caption{Hyperparameter sensitivity analysis with varying sizes of prototype banks on Weather and ETTh2 dataset. Here we use iTransformer as backbone. Left is of common pattern set, right is of rare patterns set. The look-back length is fixed to 96.}
    \label{layers}
\end{figure*}

The sizes of the Common Pattern Bank $\mathcal{B}_c$ and Rare Pattern Bank $\mathcal{B}_r$ determine the representational capacity and specialization degree of our Dual‑Prototype Bank. To analyze their impact, we set $\mathcal{B}_c \in \{32, 64, 128, 256\}$ and $\mathcal{B}_r \in \{8, 12, 24, 32\}$ respectively on the Weather and ETTh2 dataset. As shown in Figure \ref{layers}, forecasting performance remains stable across a broad range of sizes, indicating that DPAD is not highly sensitive to the size of prototype banks. A very small common bank may provide insufficient coverage of prevalent temporal dynamics, whereas a very small rare bank may limit the representation of infrequent variations. Conversely, excessively large banks provide no further gain and may increase optimization difficulty and prototype redundancy. The above observation confirms that DPAD can maintain effective role specialization and pattern coverage without requiring delicate tuning of the bank sizes.

\begin{figure*}[!t]
  \centering
    \begin{subfigure}{0.45\textwidth}
        \includegraphics[width=\columnwidth]{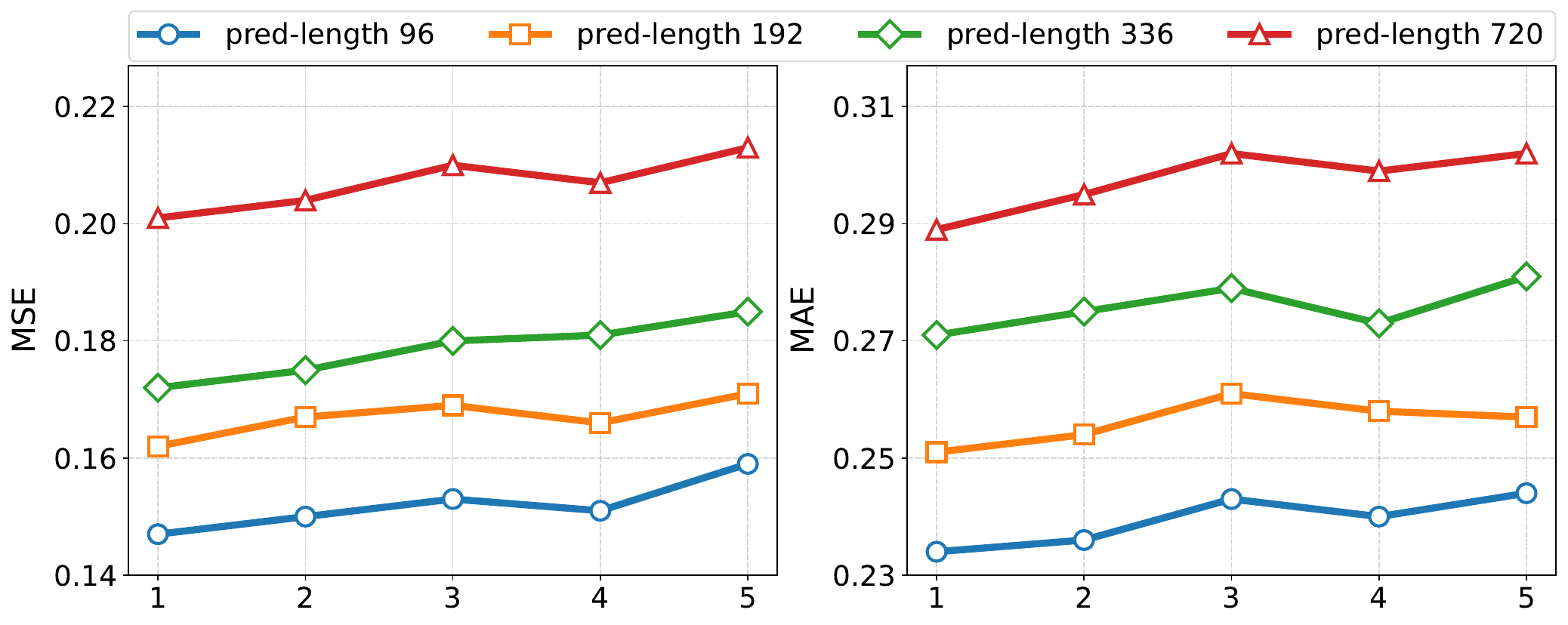}
        \caption{On ECL dataset.}
        \label{prototype11}
    \end{subfigure}
    \hspace{-0.1cm}
    \begin{subfigure}{0.45\textwidth}
        \includegraphics[width=\columnwidth]{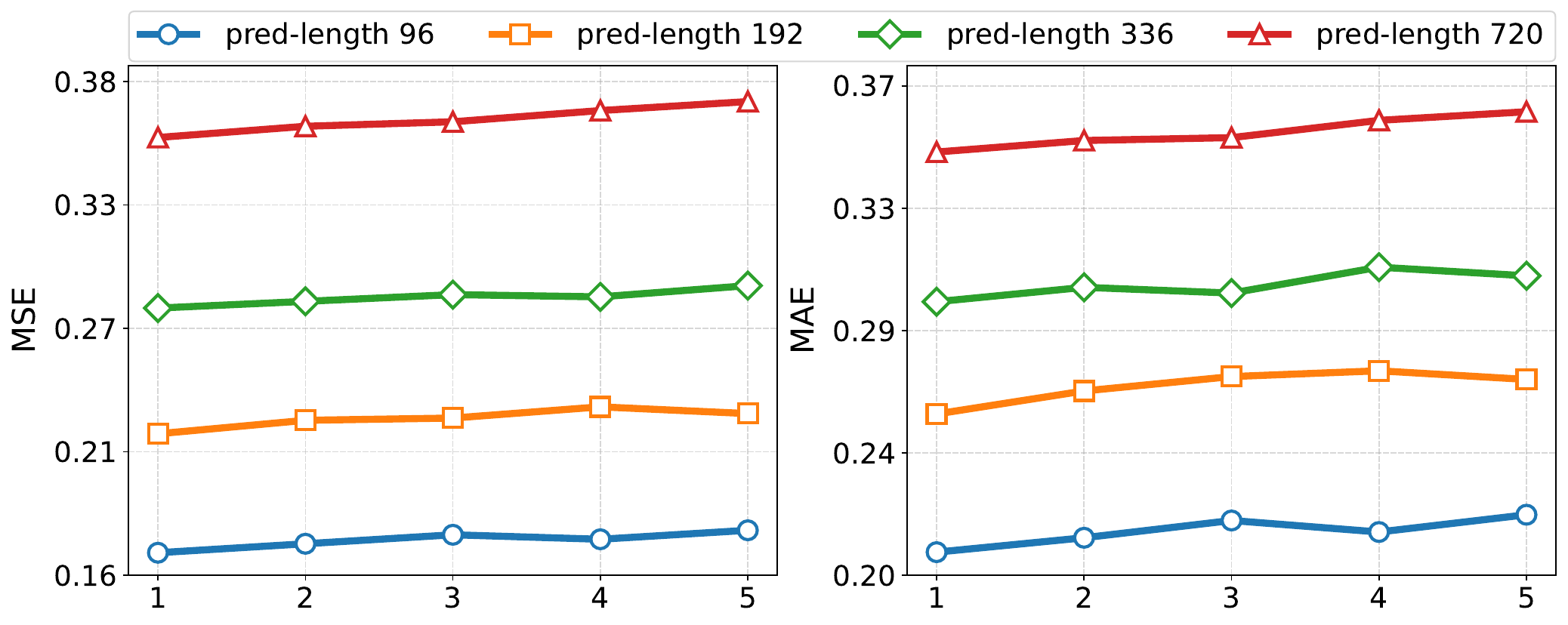}
        \caption{On Weather dataset.}
        \label{prototype21}
    \end{subfigure}
    \caption{Hyperparameter sensitivity analysis on the number of retrieved rare prototypes $K_r$ on the Electricity and Weather datasets using iTransformer as the backbone.}
    \label{rare_topk}
\end{figure*}

\begin{figure*}[!t]
  \centering
    \begin{subfigure}{0.45\textwidth}
        \includegraphics[width=\columnwidth]{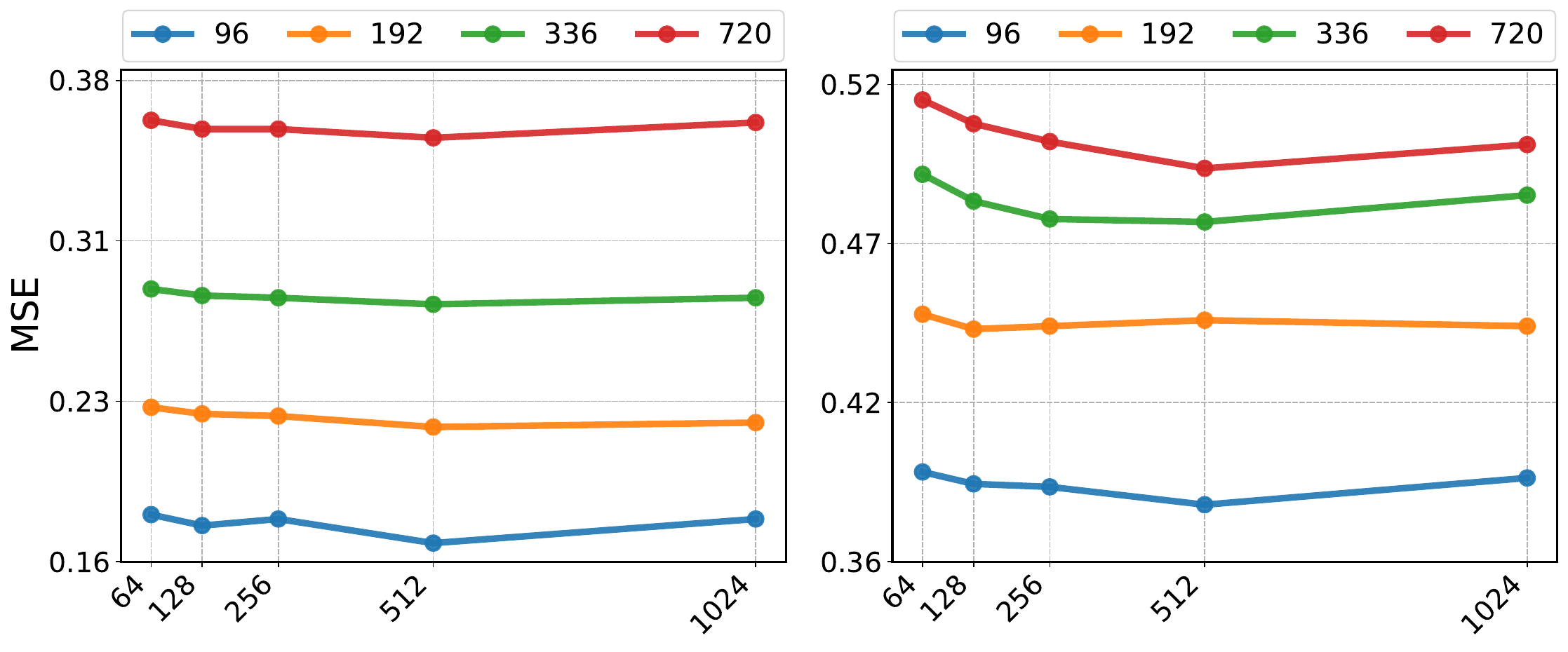}
        \caption{With iTransformer as backbone.}
        \label{fig:embed1}
    \end{subfigure}
    \hspace{-0.1cm}
    \begin{subfigure}{0.45\textwidth}
        \includegraphics[width=\columnwidth]{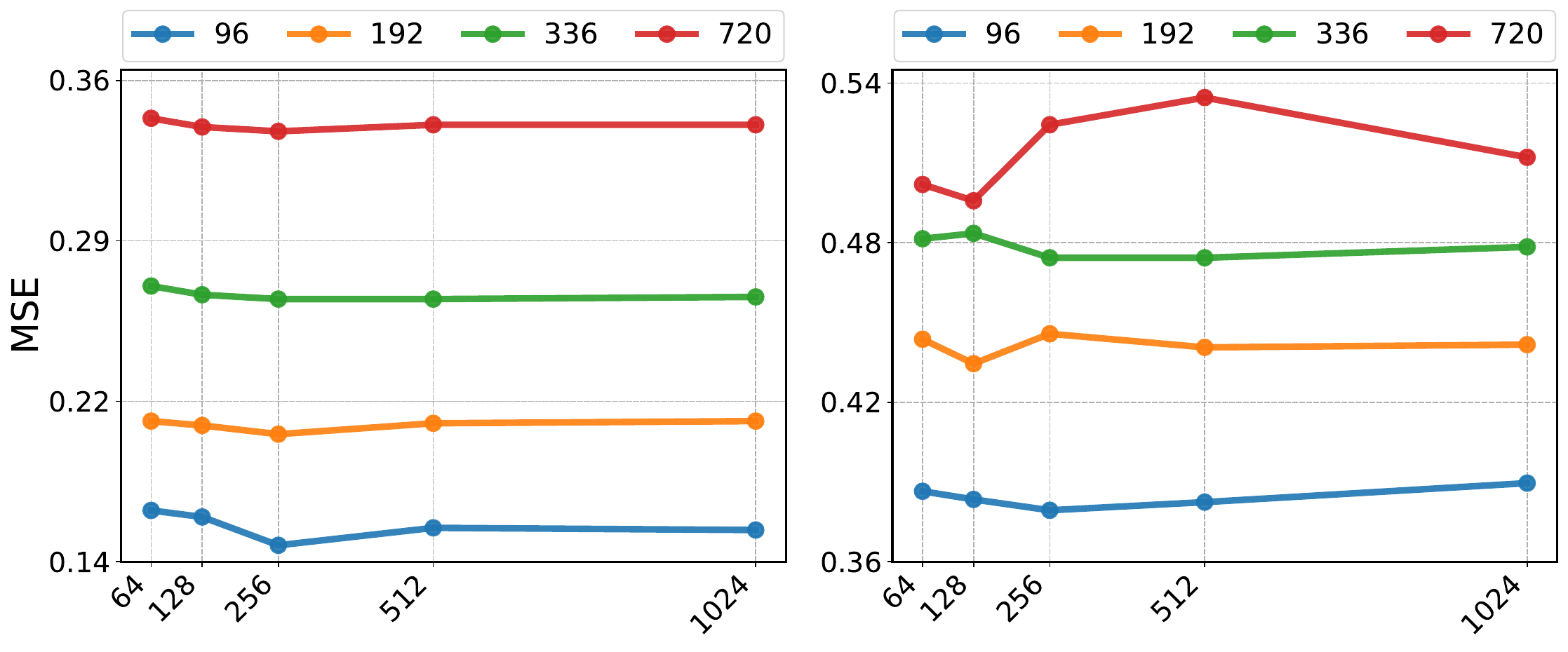}
        \caption{With TimeXer as backbone.}
        \label{fig:embed2}
    \end{subfigure}
    \begin{subfigure}{0.45\textwidth}
        \includegraphics[width=\columnwidth]{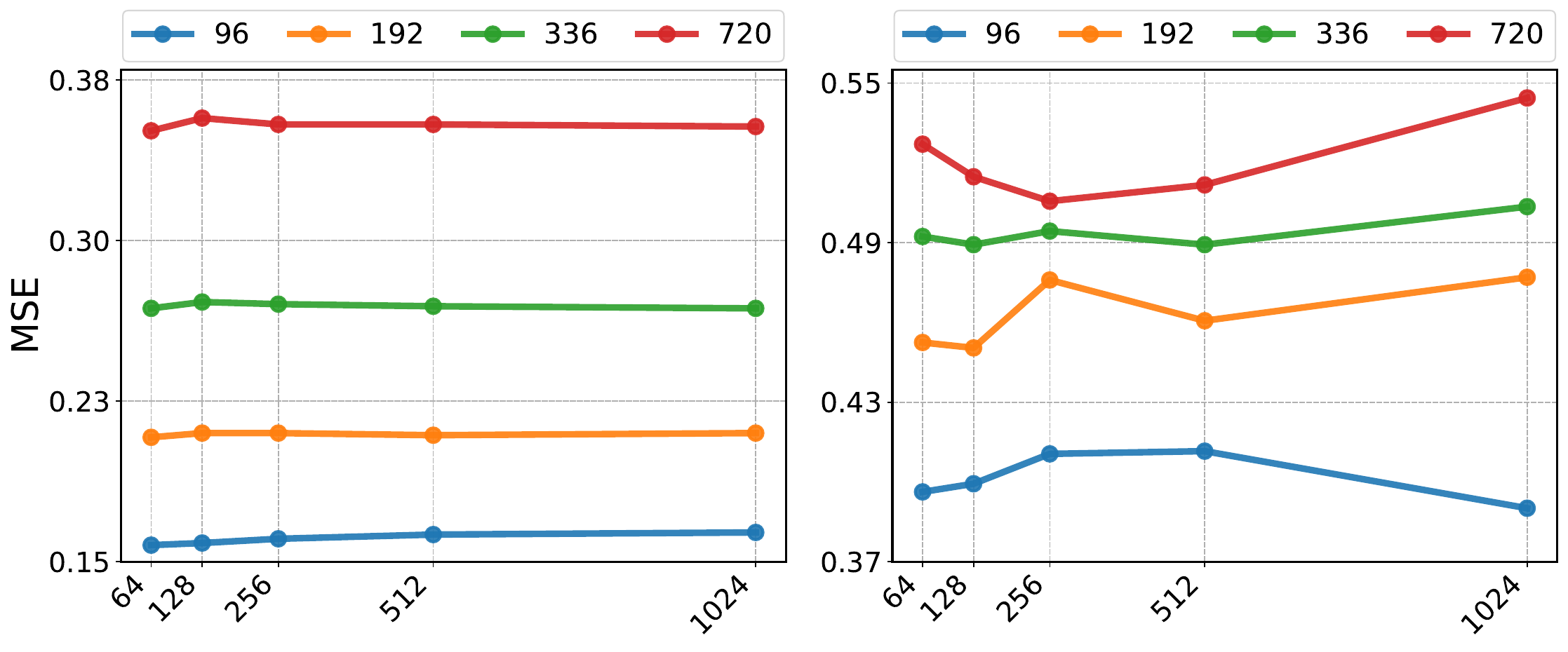}
        \caption{With TimesNet as backbone.}
        \label{fig:embed3}
    \end{subfigure}
    \hspace{-0.1cm}
    \begin{subfigure}{0.45\textwidth}
        \includegraphics[width=\columnwidth]{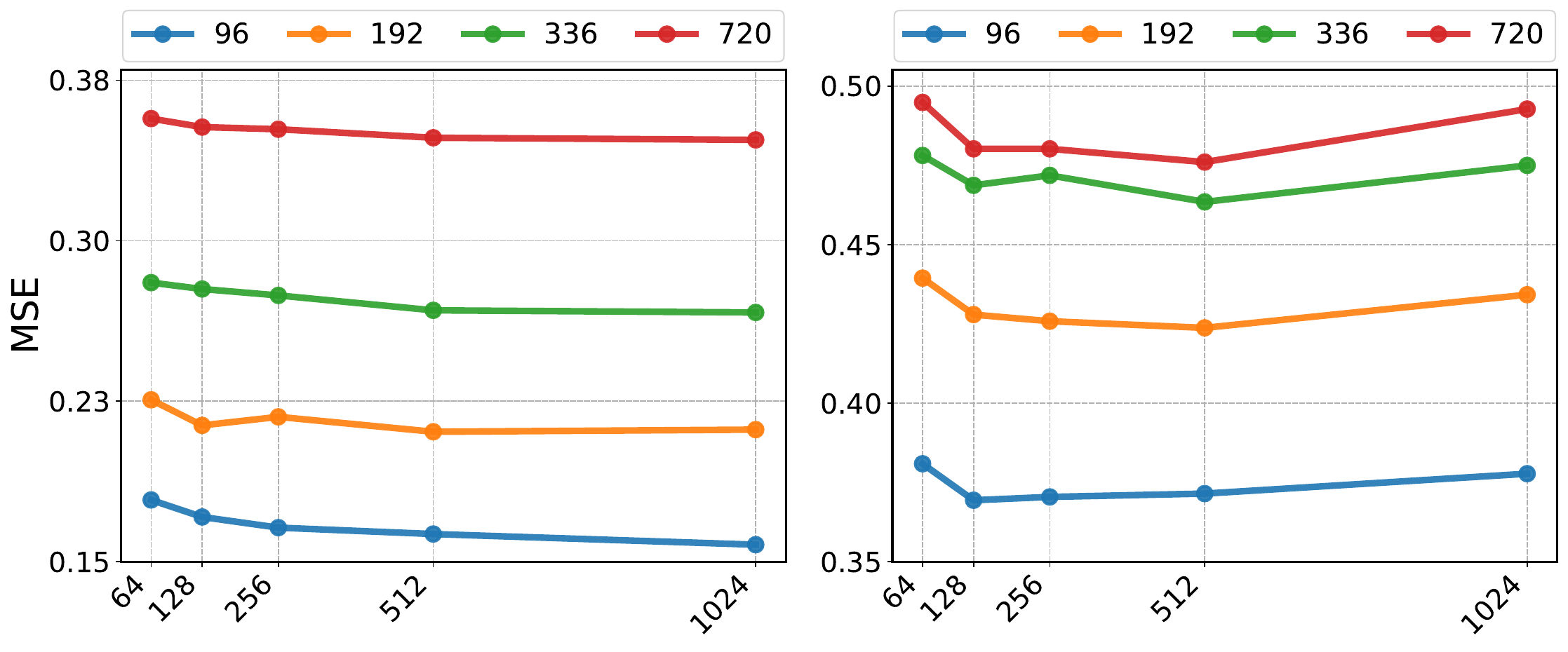}
        \caption{With TimeBridge as backbone.}
        \label{fig:embed4}
    \end{subfigure}
    \caption{Hyperparameter sensitivity analysis on embedding dimension. For each backbone, forecasting results are on the Weather (left) and ETTh1 (right) datasets. The look-back length length is fixed to 96.}
    \label{full_embed}
\end{figure*}

\subsection{Number of Retrieved Rare Prototypes}
The number of retrieved rare prototypes $K_r$ controls the sparsity of rare pattern activation in the Dual-Path Context-aware routing mechanism. A larger $K_r$ allows multiple rare prototypes to be simultaneously activated, potentially capturing co-occurring rare events, while a smaller $K_r$ enforces stricter specialization. To analyze its impact, we vary $K_r \in \{1, 2, 3, 4, 5\}$ on the Electricity and Weather datasets using iTransformer as the backbone. As shown in Figure \ref{rare_topk}, setting $K_r = 1$ achieves the best or competitive performance across both datasets. Increasing $K_r$ introduces additional rare prototype activations that may dilute the specificity of the retrieved rare information, leading to marginal degradation or no further improvement. This observation validates that a single, precisely selected rare prototype provides sufficient complementary information for infrequent patterns, and supports the design choice of enforcing sparsity in rare bank retrieval.

\subsection{Embedding Dimension}
The embedding dimension determines the capacity of the unified projection space in which the input context and role-specialized prototypes are matched. We vary $d_{\text{model}} \in \{64, 128, 256, 512, 1024\}$ on the Weather and ETTh1 dataset with four backbone models to evaluate its impact. As shown in Figure \ref{full_embed}, in most cases DPAD exhibits low sensitivity to the choice of embedding dimension. Performance does not degrade significantly even at relatively small dimensions, whereas larger dimensions provide only marginal or inconsistent gains. Based on these results, we use an embedding dimension of 128/256 in most cases. The above observation indicates that DPAD can effectively organize and selectively utilize common and rare patterns without relying on a high-dimensional latent space, facilitating its use as a model-agnostic enhancement framework.

\begin{figure*}[!t]
  \centering
    \begin{subfigure}{0.85\textwidth}
        \includegraphics[width=\textwidth]{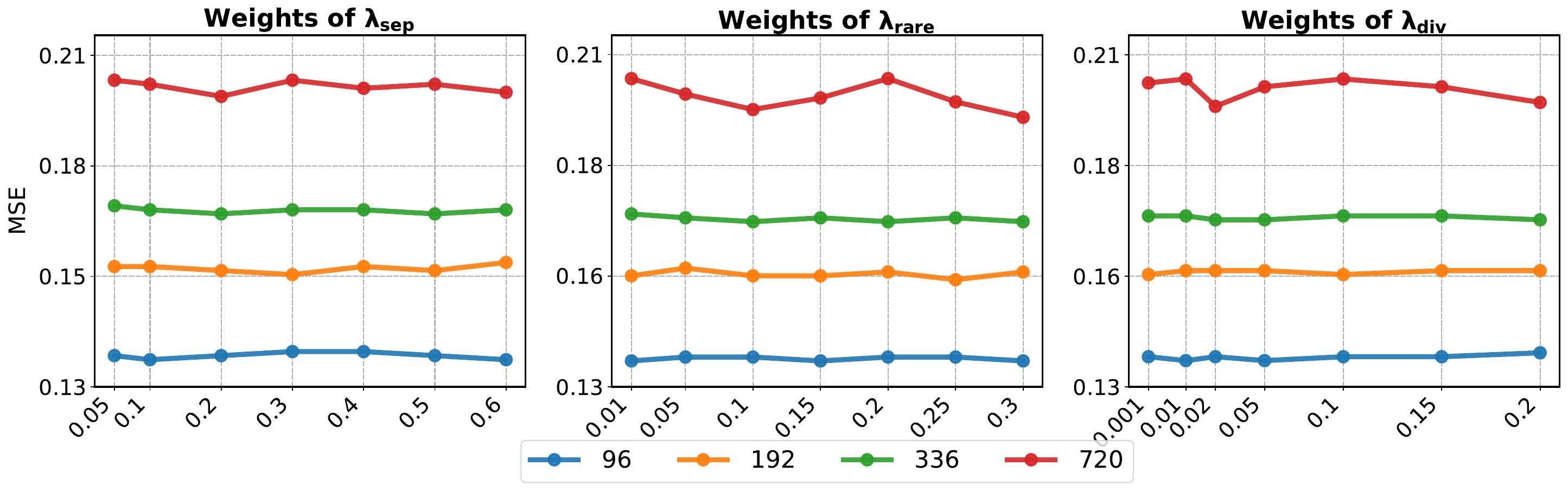}
        \caption{On ECL dataset.}
        \label{fig:loss1}
    \end{subfigure}
    \hspace{-0.1cm}
    \begin{subfigure}{0.85\textwidth}
        \includegraphics[width=\textwidth]{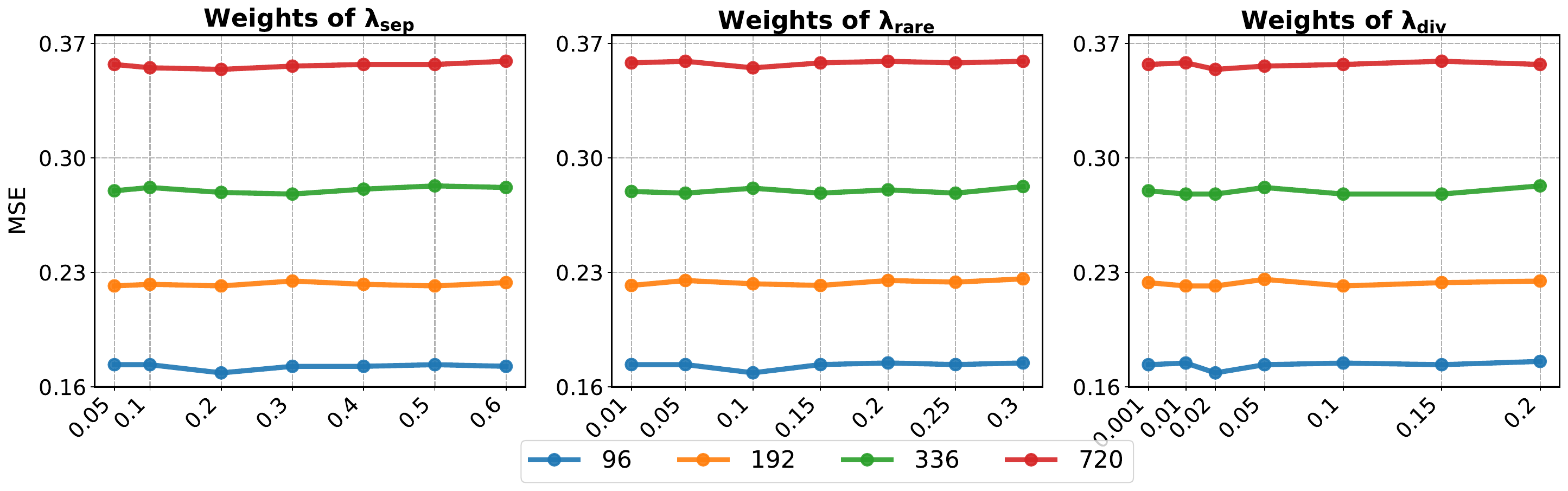}
        \caption{On Weather dataset.}
        \label{fig:loss2}
    \end{subfigure}
    \caption{Hyperparameter sensitivity analysis on weights of DGLoss. Forecasting results are on the ECL and Weather datasets, and we use TimeBridge as backbone model. The look-back length is fixed to 96.}
    \label{loss_weights}
\end{figure*}

\subsection{Weights of DGLoss}
The weights $\lambda_\text{sep}, \lambda_\text{rare}, \lambda_\text{div}$ of the DGLoss control the strengths of separation, rarity preservation, and diversity of temporal prototypes. Together, these objectives encourage the two prototype banks to specialize in their designated forecasting roles while maintaining comprehensive pattern coverage. To analyze their individual effects, we vary them in $0.001 - 0.6$ on electricity and weather datasets. As shown in Figure \ref{loss_weights}, the forecasting performance remains stable across a wide range of weight values. Specifically, the MSE exhibits only minor fluctuations when the weights are varied from $0.001 - 0.6$, indicating that the proposed DGLoss is not highly sensitive to the exact choice of these hyperparameters, which simplifies the deployment of DPAD in practice.

\begin{figure*}[!t]
    \centering
    \begin{subfigure}{\textwidth}
        \begin{subfigure}{0.245\textwidth}
            \includegraphics[width=\textwidth]{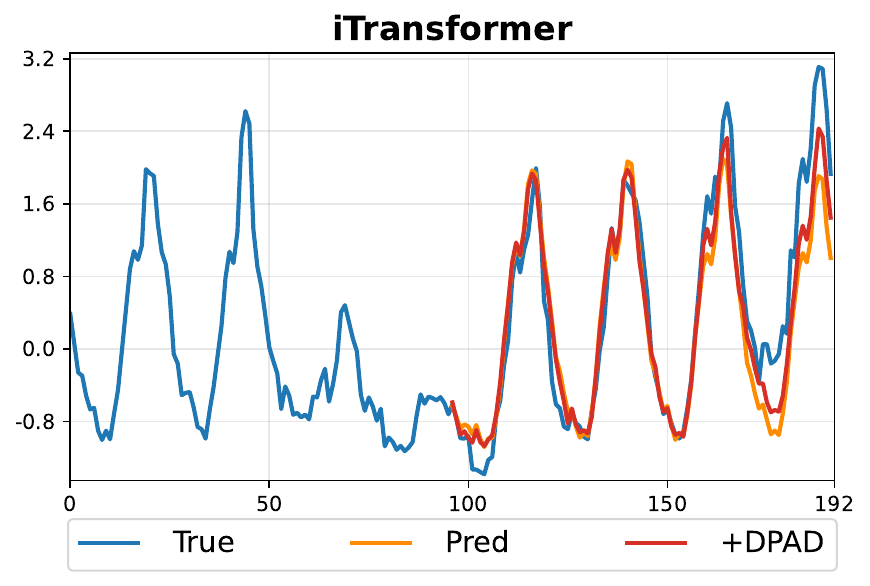}
        \end{subfigure}
        \hspace{-2pt}
        \begin{subfigure}{0.245\textwidth}
            \includegraphics[width=\textwidth]{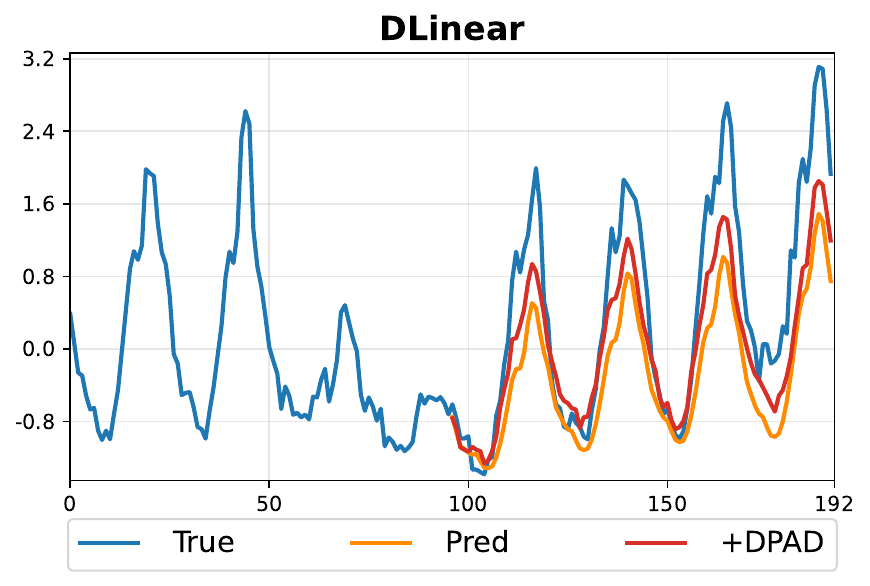}
        \end{subfigure}
        \hspace{-2pt}
        \begin{subfigure}{0.245\textwidth}
            \includegraphics[width=\textwidth]{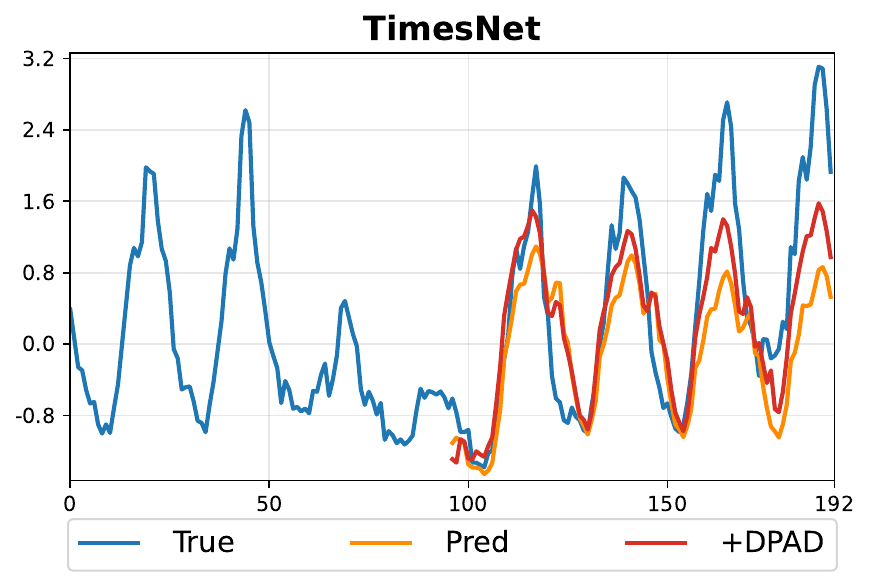}
        \end{subfigure}
        \hspace{-2pt}
        \begin{subfigure}{0.245\textwidth}
            \includegraphics[width=\textwidth]{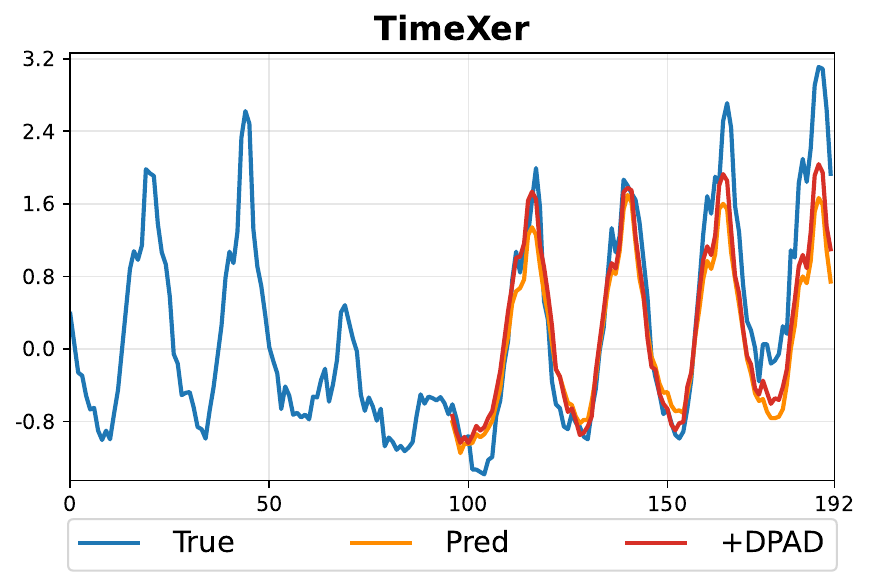}
        \end{subfigure}
    \caption{Visualization on Electricity dataset.}
    \end{subfigure}
    \\
    \begin{subfigure}{\textwidth}
        \begin{subfigure}{0.245\textwidth}
            \includegraphics[width=\textwidth]{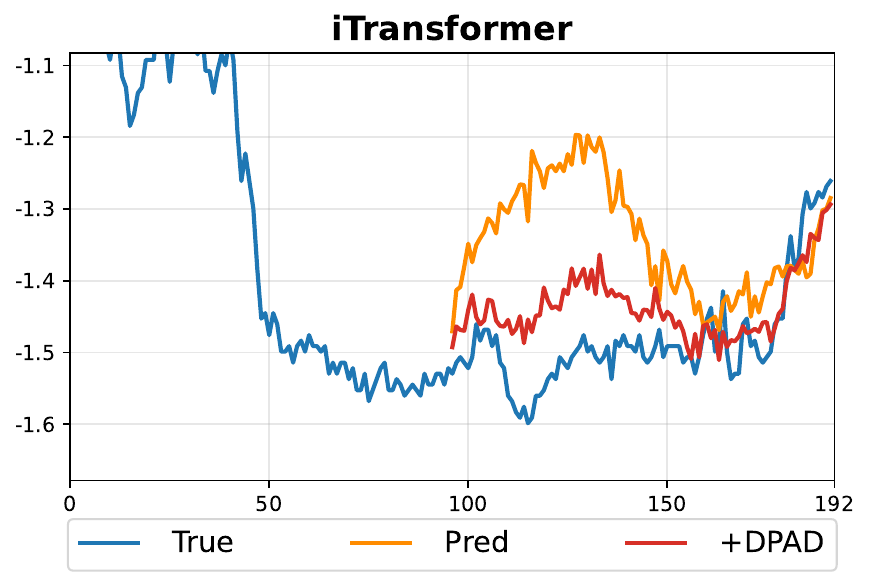}
        \end{subfigure}
        \hspace{-2pt}
        \begin{subfigure}{0.245\textwidth}
            \includegraphics[width=\textwidth]{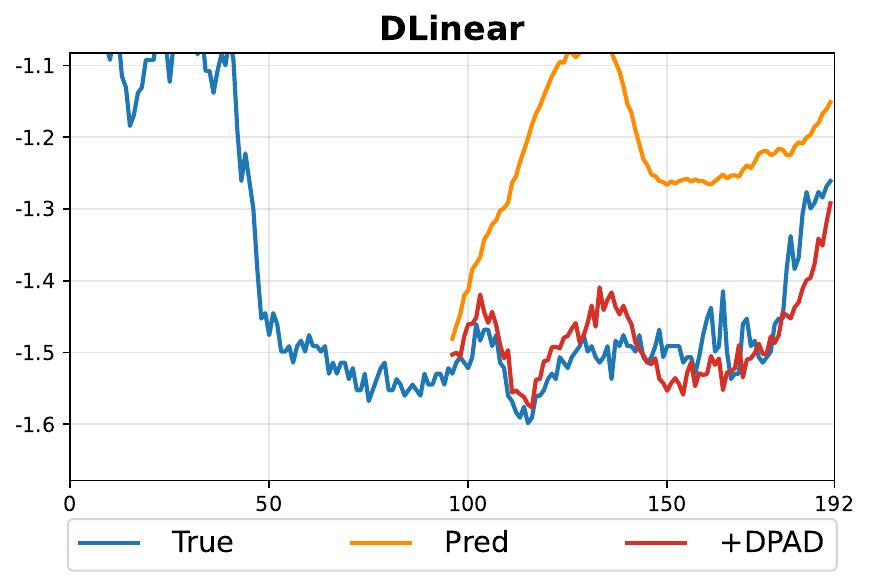}
        \end{subfigure}
        \hspace{-2pt}
        \begin{subfigure}{0.245\textwidth}
            \includegraphics[width=\textwidth]{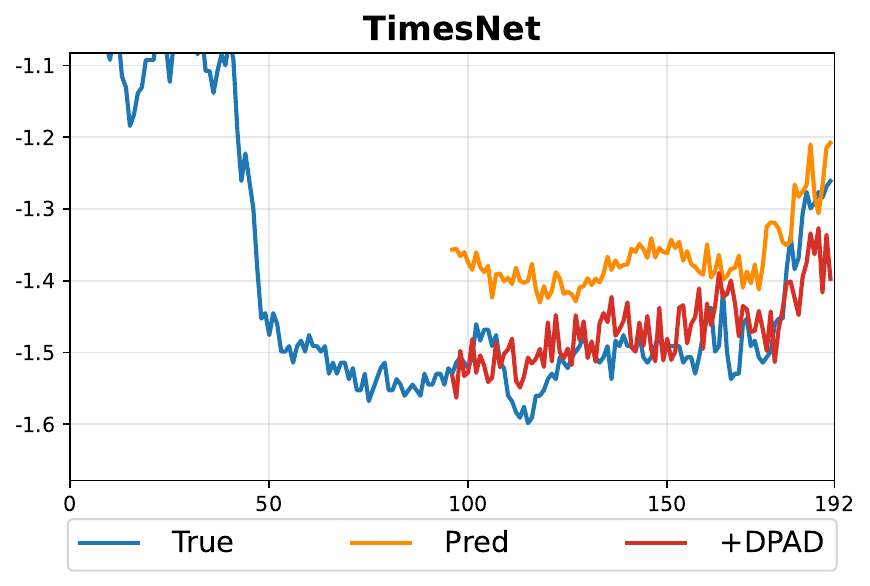}
        \end{subfigure}
        \hspace{-2pt}
        \begin{subfigure}{0.245\textwidth}
            \includegraphics[width=\textwidth]{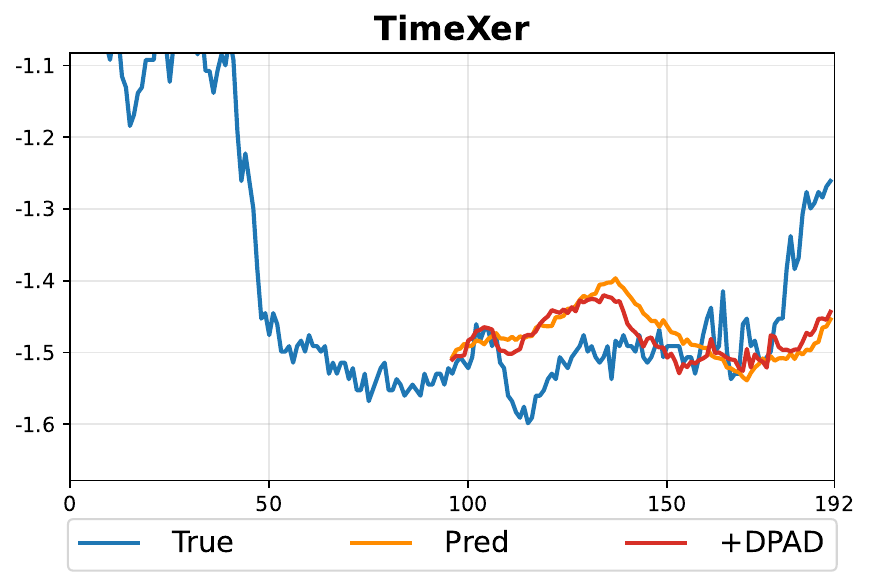}
        \end{subfigure}
    \caption{Visualization on ETTm1 dataset.}
    \end{subfigure}
    \\
    \begin{subfigure}{\textwidth}
        \begin{subfigure}{0.245\textwidth}
            \includegraphics[width=\textwidth]{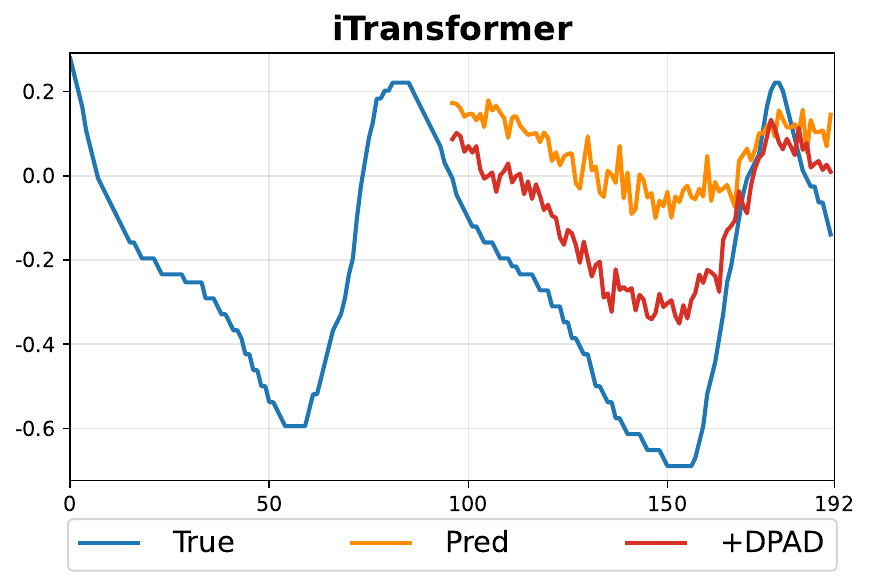}
        \end{subfigure}
        \hspace{-2pt}
        \begin{subfigure}{0.245\textwidth}
            \includegraphics[width=\textwidth]{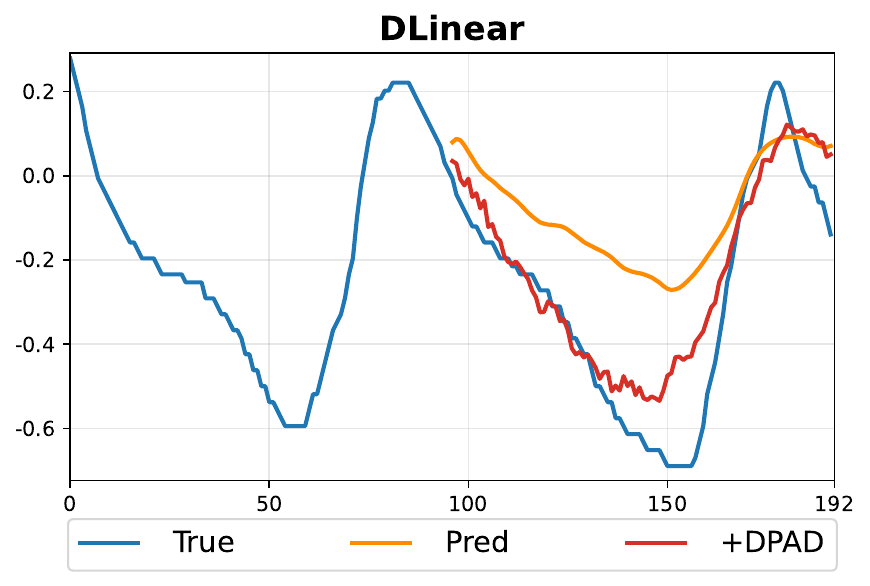}
        \end{subfigure}
        \hspace{-2pt}
        \begin{subfigure}{0.245\textwidth}
            \includegraphics[width=\textwidth]{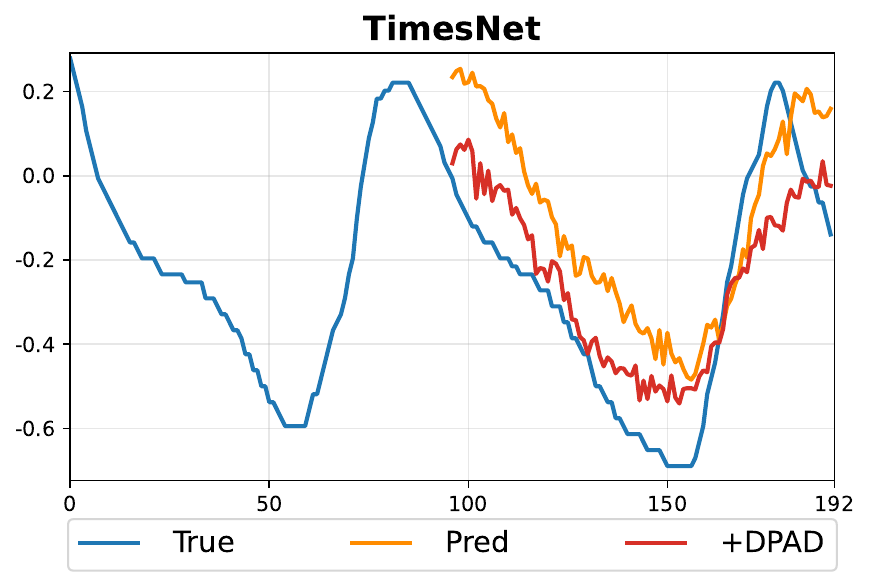}
        \end{subfigure}
        \hspace{-2pt}
        \begin{subfigure}{0.245\textwidth}
            \includegraphics[width=\textwidth]{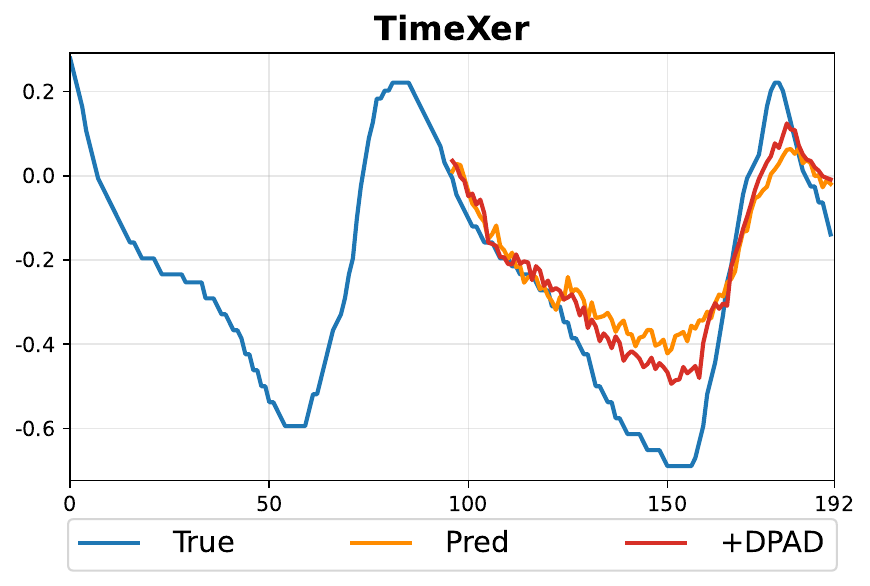}
        \end{subfigure}
    \caption{Visualization on ETTm2 dataset.}
    \end{subfigure}
    \\
    \begin{subfigure}{\textwidth}
        \begin{subfigure}{0.245\textwidth}
            \includegraphics[width=\textwidth]{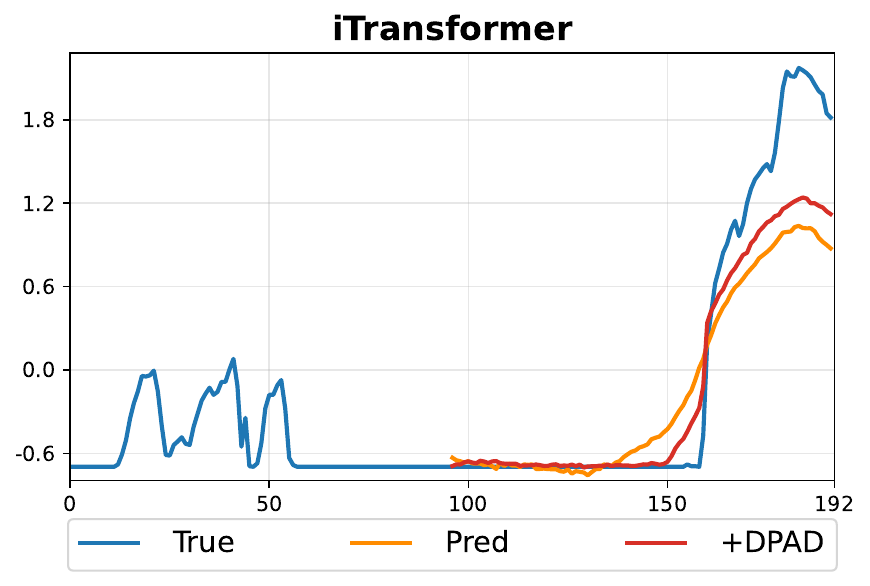}
        \end{subfigure}
        \hspace{-2pt}
        \begin{subfigure}{0.245\textwidth}
            \includegraphics[width=\textwidth]{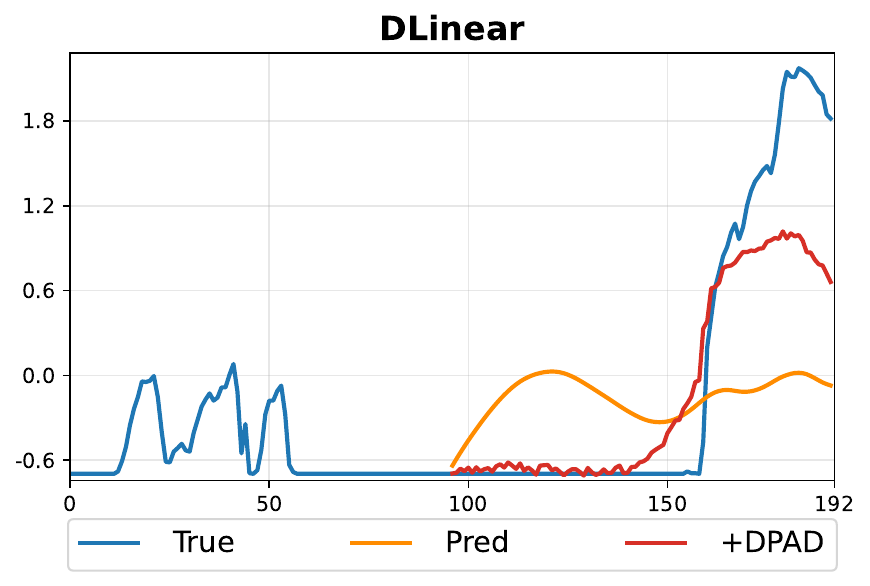}
        \end{subfigure}
        \hspace{-2pt}
        \begin{subfigure}{0.245\textwidth}
            \includegraphics[width=\textwidth]{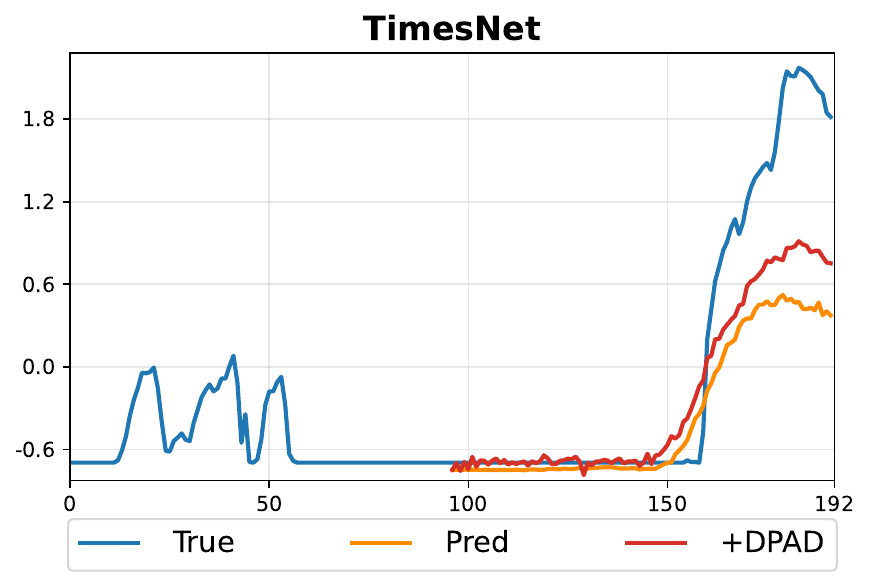}
        \end{subfigure}
        \hspace{-2pt}
        \begin{subfigure}{0.245\textwidth}
            \includegraphics[width=\textwidth]{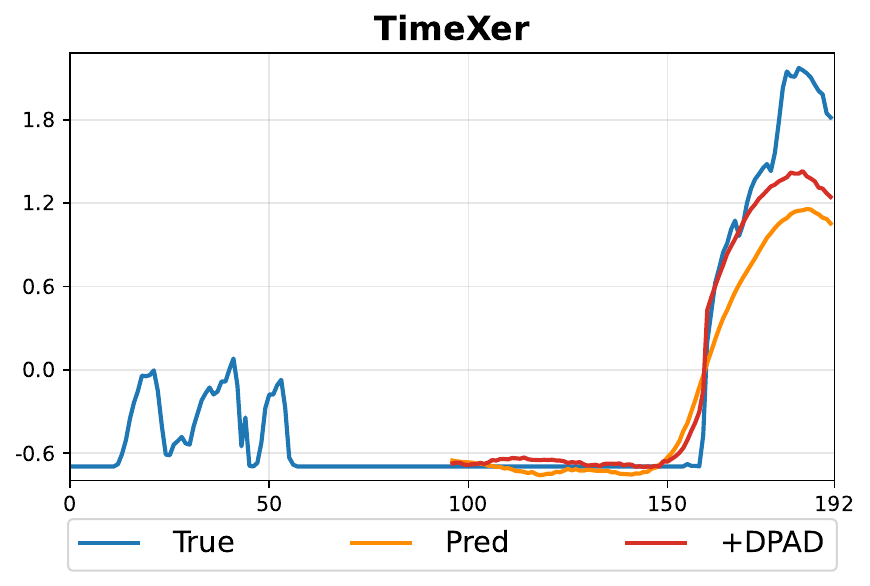}
        \end{subfigure}
    \caption{Visualization on Solar dataset.}
    \end{subfigure}
    \caption{Forecasting visualization for backbone models with and without DPAD.}
    \label{forecast_visual}
\end{figure*}

\section{Forecasting Visualization}
To provide a more intuitive and comprehensive view of the improvements brought by our DPAD framework, we present forecasting visualizations comparing the predictions of backbone models with and without DPAD across four representative datasets. As shown in Figure \ref{forecast_visual}, the predictions with DPAD generally follow the ground truth more closely across different temporal dynamics and model architectures. In particular, DPAD improves the modeling of both recurring structures and context-specific variations rather than focusing exclusively on a single pattern category. These qualitative results further demonstrate that role-aware pattern organization and context-aware prototype utilization can provide complementary forecasting information across diverse backbones and datasets.

\section{Full Results}
\subsection{Main Experiments}
We evaluate DPAD on seven real-world TSF benchmarks spanning diverse domains. Table \ref{full_results1} shows the full results of long-term forecasting tasks on ETT (ETTh1, ETTh2, ETTm1, ETTm2), Electricity, Exchange, Solar, Weather, and Traffic datasets. Table \ref{full_results2} contains the full results of short-term forecasting tasks on PEMS (PEMS03, PEMS04, PEMS07, PEMS08) datasets. From the full results, we can see that DPAD achieves consistent improvements on most backbones and datasets. These results support the effectiveness of organizing heterogeneous temporal patterns according to their forecasting roles and selectively utilizing them based on the input context. 

\subsection{Ablation Results}
Table \ref{full_ablation} shows the full results of ablation studies discussed in the main text. From full results, we can see that each component in our DPAD framework plays an indispensable role. These results demonstrate that DDP, DPC, and DGLoss play complementary roles in enabling role-aware pattern organization and context-aware pattern utilization.

\begin{table*}[!t]
\resizebox{\textwidth}{!}{
\begin{tabular}{c|c|cc|cc|cc|cc|cc|cc|cc|cc|cc|cc}
    \toprule
\multicolumn{2}{c|}{\multirow{2}{*}{Methods}} & \multicolumn{4}{c|}{iTransformer} & \multicolumn{4}{c|}{DLinear} & \multicolumn{4}{c|}{TimesNet} & \multicolumn{4}{c|}{TimeXer} & \multicolumn{4}{c}{TimeBridge} \\
\cmidrule(lr){3-22}
 \multicolumn{2}{c|}{}& \multicolumn{2}{c|}{Ori} & \multicolumn{2}{c|}{+DPAD} & \multicolumn{2}{c|}{Ori} & \multicolumn{2}{c|}{+DPAD} & \multicolumn{2}{c|}{Ori} & \multicolumn{2}{c|}{+DPAD} & \multicolumn{2}{c|}{Ori} & \multicolumn{2}{c|}{+DPAD} & \multicolumn{2}{c|}{Ori} & \multicolumn{2}{c}{+DPAD} \\     \midrule
\multicolumn{2}{c|}{Metric} & MSE & MAE & MSE & MAE & MSE & MAE & MSE & MAE & MSE & MAE & MSE & MAE & MSE & MAE & MSE & MAE & MSE & MAE & MSE & MAE \\ \midrule
 \multirow[c]{5}{*}{\rotatebox{90}{ETTh1}}& {96} & 0.387  & 0.405  & \textbf{0.383 } & \textbf{0.401 } & 0.397  & 0.412  & \textbf{0.388 } & \textbf{0.403 } & \textbf{0.415 } & 0.429  & 0.419  & \textbf{0.423 } & 0.386  & 0.404  & \textbf{0.379 } & \textbf{0.398 } & 0.377  & 0.395  & \textbf{0.371 } & \textbf{0.388 } \\
 & {192} & 0.441  & 0.436  & \textbf{0.433 } & \textbf{0.423 } & 0.446  & 0.441  & \textbf{0.431 } & \textbf{0.432 } & 0.479  & 0.466  & \textbf{0.471 } & \textbf{0.460 } & \textbf{0.429 } & 0.435  & 0.433  & \textbf{0.430 } & 0.427  & 0.425  & \textbf{0.423 }  & \textbf{0.421 } \\
 & {336} & 0.494  & 0.463  & \textbf{0.478 } & \textbf{0.452 } & 0.489  & 0.467  & \textbf{0.478 } & \textbf{0.459 } & 0.517  & 0.482  & \textbf{0.504 } & \textbf{0.473 } & 0.484  & 0.457  & \textbf{0.472 } & \textbf{0.446 } & 0.477  & 0.456  & \textbf{0.461 } & \textbf{0.441 } \\
 & {720} & \textbf{0.488 } & 0.483  & 0.492  & \textbf{0.482 } & 0.513  & 0.511  & \textbf{0.501 } & \textbf{0.506 } & \textbf{0.505 } & \textbf{0.490 } & 0.512  & 0.496  & 0.544  & 0.513  & \textbf{0.483 } & \textbf{0.493 } & 0.521  & 0.498  & \textbf{0.473 } & \textbf{0.474 } \\\cmidrule{2-22} 
 & {Avg} & 0.452  & 0.446  & \textbf{0.446 } & \textbf{0.439 } & 0.461  & 0.457  & \textbf{0.449 } & \textbf{0.450 } & 0.479  & 0.466  & \textbf{0.476 } & \textbf{0.463 } & 0.460  & 0.452  & \textbf{0.442 } & \textbf{0.441 } & 0.450  & 0.443  & \textbf{0.432 } & \textbf{0.431 } \\\midrule
  \multirow[c]{5}{*}{\rotatebox{90}{ETTh2}}& {96} & 0.301  & 0.350  & \textbf{0.293 } & \textbf{0.345 } & 0.341  & 0.394  & \textbf{0.332 } & \textbf{0.381 } & 0.316  & 0.358  & \textbf{0.309 } & 0.358  & 0.284  & 0.337  & \textbf{0.276 } & \textbf{0.331 } & 0.294  & 0.345  & \textbf{0.287 } & \textbf{0.333 } \\
 & {192} & 0.380  & 0.399  & \textbf{0.374 } & \textbf{0.393 } & 0.482  & 0.479  & \textbf{0.466 } & \textbf{0.471 } & 0.415  & 0.414  & \textbf{0.397 } & \textbf{0.407 } & 0.366  & \textbf{0.391 } & 0.366  & 0.392  & 0.377  & 0.395  & \textbf{0.359 } & \textbf{0.380 } \\
 & {336} & 0.423  & 0.431  & \textbf{0.416 } & \textbf{0.425 } & 0.591  & 0.541  & \textbf{0.535 } & \textbf{0.512 } & 0.452  & 0.448  & \textbf{0.433 } & \textbf{0.436 } & 0.438  & 0.438  & \textbf{0.424 } & \textbf{0.429 } & 0.424  & 0.433  & \textbf{0.408 } & \textbf{0.418 } \\
 & {720} & 0.431  & 0.447  & \textbf{0.422 } & \textbf{0.435 } & 0.839  & 0.661  & \textbf{0.785 } & \textbf{0.638 } & 0.461  & 0.463  & \textbf{0.448 } & \textbf{0.455 } & \textbf{0.407 } & 0.449  & 0.411  & \textbf{0.432 } & 0.423  & 0.442  & \textbf{0.417 } & \textbf{0.435 } \\\cmidrule{2-22} 
 & {Avg} & 0.383  & 0.406  & \textbf{0.376 } & \textbf{0.399 } & 0.563  & 0.518  & \textbf{0.529 } & \textbf{0.500 } & 0.411  & 0.420  & \textbf{0.396 } & \textbf{0.414 } & 0.373  & 0.403  & \textbf{0.369 } & \textbf{0.396 } & 0.379  & 0.403  & \textbf{0.367 } & \textbf{0.391 } \\\midrule
  \multirow[c]{5}{*}{\rotatebox{90}{ETTm1}}& {96} & 0.341  & 0.376  & \textbf{0.332 } & \textbf{0.369 } & 0.346  & 0.374  & \textbf{0.334 } & \textbf{0.369 } & 0.336  & 0.375  & \textbf{0.329 } & \textbf{0.371 } & \textbf{0.318 } & \textbf{0.356 } & 0.320  & 0.358  & 0.324  & 0.362  & \textbf{0.311 } & \textbf{0.345 } \\
 & {192} & 0.382  & 0.396  & \textbf{0.377 } & \textbf{0.389 } & 0.382  & \textbf{0.391 } & \textbf{0.375 } & 0.392  & 0.377  & 0.395  & \textbf{0.376 } & \textbf{0.393 } & 0.373  & 0.389  & \textbf{0.366 } & \textbf{0.383 } & 0.365  & 0.384  & \textbf{0.361 } & \textbf{0.375 } \\
 & {336} & 0.420  & 0.421  & \textbf{0.411 } & \textbf{0.412 } & 0.415  & 0.415  & \textbf{0.405 } & 0.415  & 0.418  & 0.420  & \textbf{0.405 } & \textbf{0.409 } & 0.412  & \textbf{0.387 } & \textbf{0.407 } & 0.394  & 0.398  & 0.407  & \textbf{0.392 } & \textbf{0.397 } \\
 & {720} & 0.487  & 0.456  & \textbf{0.484 } & \textbf{0.451 } & 0.473  & 0.451  & \textbf{0.465 } & \textbf{0.443 } & 0.541  & 0.481  & \textbf{0.493 } & \textbf{0.464 } & 0.460  & 0.450  & \textbf{0.451 } & \textbf{0.441 } & 0.462  & 0.444  & \textbf{0.455 } & \textbf{0.431 } \\\cmidrule{2-22} 
 & {Avg} & 0.407  & 0.412  & \textbf{0.401 } & \textbf{0.405 } & 0.404  & 0.407  & \textbf{0.394 } & \textbf{0.404 } & 0.418  & 0.417  & \textbf{0.400 } & \textbf{0.409 } & 0.390  & 0.395 & \textbf{0.386 } & \textbf{0.394 }  & 0.387  & 0.399  & \textbf{0.379 } & \textbf{0.387 }  \\\midrule
  \multirow[c]{5}{*}{\rotatebox{90}{ETTm2}}& {96} & 0.186  & 0.272  & \textbf{0.179 } & \textbf{0.263 } & 0.193  & \textbf{0.293 } & \textbf{0.190 } & 0.295  & 0.188  & 0.268  & \textbf{0.181 } & \textbf{0.264 } & 0.172  & 0.254  & \textbf{0.165 } & \textbf{0.251 } & 0.177  & 0.260  & \textbf{0.176 } & \textbf{0.254 }  \\
 & {192} & 0.252  & 0.312  & \textbf{0.245 } & \textbf{0.306 } & 0.284  & 0.361  & \textbf{0.273 } & \textbf{0.353 } & 0.250  & 0.306  & \textbf{0.243 } & \textbf{0.304 } & 0.241  & 0.302  & \textbf{0.238 } & \textbf{0.300 } & 0.243  & 0.303  & \textbf{0.242 } & \textbf{0.298 } \\
 & {336} & 0.315  & 0.351  & \textbf{0.309 } & \textbf{0.344 } & 0.382  & 0.429  & \textbf{0.367 } & \textbf{0.415 } & 0.306  & 0.341  & \textbf{0.298 } & \textbf{0.337 } & \textbf{0.301 } & 0.340  & 0.397  & \textbf{0.331 } & \textbf{0.303 } & 0.342  & 0.304  & \textbf{0.338 }  \\
 & {720} & 0.415  & 0.408  & \textbf{0.409 } & \textbf{0.405 } & 0.558  & 0.525  & \textbf{0.508 } & \textbf{0.501 } & 0.420  & 0.405  & \textbf{0.402 } & \textbf{0.398 } & \textbf{0.394 } & 0.395  & 0.396  & \textbf{0.393 } & \textbf{0.401 } & 0.399  & 0.409  & \textbf{0.397 } \\\cmidrule{2-22} 
 & {Avg} & 0.292  & 0.335  & \textbf{0.285 } & \textbf{0.329 } & 0.354  & 0.402  & \textbf{0.334 } & \textbf{0.391 } & 0.291  & 0.330  & \textbf{0.281 } & \textbf{0.325 } & \textbf{0.277 } & 0.322  & 0.299  & \textbf{0.318 } & \textbf{0.281 } & 0.326  & 0.282  & \textbf{0.321 } \\\midrule
  \multirow[c]{5}{*}{\rotatebox{90}{Electricity}} & {96} & 0.148  & 0.241  & \textbf{0.147 } & \textbf{0.234 } & 0.211  & 0.302  & \textbf{0.202 } & \textbf{0.293 } & 0.163  & \textbf{0.267 } & \textbf{0.159 } & 0.269  & 0.241  & 0.244  & \textbf{0.141 } & \textbf{0.241 } & 0.143  & 0.239  & \textbf{0.135 } & \textbf{0.228 } \\
 & {192} & 0.167  & \textbf{0.248 } & \textbf{0.162 } & 0.251  & 0.211  & 0.305  & \textbf{0.203 } & \textbf{0.293 } & 0.184  & 0.284  & \textbf{0.178 } & \textbf{0.281 } & 0.159  & 0.260  & \textbf{0.153 } & \textbf{0.257 } & 0.161  & 0.256  & \textbf{0.154 } & \textbf{0.247 } \\
 & {336}   & 0.179  & 0.271  & \textbf{0.172 } & \textbf{0.271 }  & 0.223  & 0.319  & \textbf{0.210 } & \textbf{0.304 } & \textbf{0.196 } & \textbf{0.297 } & 0.205  & 0.303  & 0.177  & 0.276  & 0.177  & \textbf{0.271 } & 0.176  & 0.272  & \textbf{0.172 } & \textbf{0.264 } \\
 & {720} & 0.208  & 0.298  & \textbf{0.201 } & \textbf{0.289 } & 0.258  & 0.351  & \textbf{0.243 } & \textbf{0.332 } & \textbf{0.232 } & \textbf{0.325 } & 0.243  & 0.331  & 0.229  & 0.321  & \textbf{0.210 } & \textbf{0.306 } & 0.204  & 0.297  & \textbf{0.201 } & \textbf{0.288 } \\\cmidrule{2-22} 
 & {Avg} & 0.175  & 0.264  & \textbf{0.170 } & \textbf{0.261 } & 0.225  & 0.319  & \textbf{0.214 } & \textbf{0.305 } & \textbf{0.193 } & \textbf{0.293 } & 0.196  & 0.296  & 0.201  & 0.275  & \textbf{0.170 } & \textbf{0.268 } & 0.171  & 0.266  & \textbf{0.165 } & \textbf{0.256 } \\\midrule
 \multirow[c]{5}{*}{\rotatebox{90}{Exchange}} & {96} & 0.088  & 0.208  & \textbf{0.081 } & \textbf{0.202 } & 0.098  & 0.233  & \textbf{0.095 } & \textbf{0.228 } & 0.115  & 0.242  & \textbf{0.097 } & \textbf{0.224 } & 0.094  & 0.214  & \textbf{0.086 } & \textbf{0.205 } & 0.113  & 0.236  & \textbf{0.084 } & \textbf{0.201 } \\
 & {192} & 0.180  & 0.303  & \textbf{0.175 } & \textbf{0.299 } & 0.186  & 0.325  & \textbf{0.182 } & \textbf{0.319 } & 0.216  & 0.333  & \textbf{0.186 } & \textbf{0.310 } & 0.182  & 0.303  & \textbf{0.179 } & \textbf{0.301 } & 0.200  & 0.327  & \textbf{0.177 } & \textbf{0.297 } \\
 & {336} & 0.331  & 0.418  & \textbf{0.326 } & \textbf{0.414 } & \textbf{0.325 } & \textbf{0.434 } & 0.329  & 0.442  & 0.375  & 0.444  & \textbf{0.360 } & \textbf{0.434 } & 0.384  & 0.448  & \textbf{0.350 } & \textbf{0.426 } & 0.364  & 0.440  & \textbf{0.339 } & \textbf{0.421 } \\
 & {720} & 0.848  & 0.695  & \textbf{0.835 } & \textbf{0.692 } & \textbf{0.746 } & \textbf{0.663 } & 0.772  & 0.678  & 1.012  & 0.765  & \textbf{0.893 }  & \textbf{0.720 } & 0.932  & 0.724  & \textbf{0.887 } & \textbf{0.706 } & 0.992  & 0.760  & \textbf{0.847 } & \textbf{0.691 }  \\\cmidrule{2-22} 
& {Avg} & 0.361  & 0.406  & \textbf{0.354 } & \textbf{0.401 } & \textbf{0.338 } & \textbf{0.413 } & 0.345  & 0.417  & 0.430 & 0.446  & \textbf{0.384 } & \textbf{0.422 } & 0.398  & 0.422  & \textbf{0.375 } & \textbf{0.409 } & 0.417  & 0.440  & \textbf{0.361 } & \textbf{0.402 } \\\midrule
\multirow[c]{5}{*}{\rotatebox{90}{Solar}} & {96} & 0.207  & 0.237  & \textbf{0.199 } & \textbf{0.221 } & 0.290  & 0.378  & \textbf{0.240 } & \textbf{0.299 } & 0.223  & 0.256  & \textbf{0.211 } & \textbf{0.249 } & 0.198  & 0.244  & \textbf{0.191 } & \textbf{0.235 } & 0.204  & 0.245  & \textbf{0.196 } & \textbf{0.212 } \\
 & {192} & 0.242  & 0.264  & \textbf{0.237 } & \textbf{0.254 } & 0.320  & 0.398  & \textbf{0.263 } & \textbf{0.312 } & 0.262  & 0.272  & \textbf{0.241 } & \textbf{0.267 } & 0.226  & 0.270  & \textbf{0.221 } & \textbf{0.262 } & 0.237  & 0.269  & \textbf{0.228 } & \textbf{0.236 } \\
 & {336} & 0.251  & 0.277  & \textbf{0.249 }  & \textbf{0.273 }  & 0.353  & 0.415  & \textbf{0.279 } & \textbf{0.314 } & 0.287  & 0.299  & \textbf{0.257 } & \textbf{0.291 } & 0.239  & 0.281  & \textbf{0.231 } & \textbf{0.275 } & 0.251  & 0.283  & \textbf{0.247 } & \textbf{0.256 } \\
 & {720} & 0.251  & 0.278  & \textbf{0.250 }  & \textbf{0.276 }  & 0.357  & 0.413  & \textbf{0.276 } & \textbf{0.310 } & 0.298  & 0.318  & \textbf{0.267 } & \textbf{0.298 } & 0.242  & 0.282  & \textbf{0.238 } & \textbf{0.277 } & \textbf{0.253 } & 0.284  & 0.255  & \textbf{0.260 } \\\cmidrule{2-22} 
& {Avg} & 0.237  & 0.264  & \textbf{0.233 } & \textbf{0.256 } & 0.330  & 0.401  & \textbf{0.264 } & \textbf{0.308 } & 0.267  & 0.286  & \textbf{0.244 } & \textbf{0.276 } & 0.226  & 0.269  & \textbf{0.220 } & \textbf{0.262 } & 0.236  & 0.270  & \textbf{0.231 } & \textbf{0.241 } \\\midrule
\multirow[c]{5}{*}{\rotatebox{90}{Traffic}} & {96} & 0.393  & 0.269  & \textbf{0.392 }  & \textbf{0.268 }  & 0.712  & 0.438  & \textbf{0.647 } & \textbf{0.402 } & 0.593  & 0.317  & \textbf{0.471 } & \textbf{0.295 } & 0.428  & 0.271  & \textbf{0.415 } & \textbf{0.263 } & 0.375  & 0.270  & \textbf{0.371 } & \textbf{0.243 } \\
 & {192} & 0.412  & 0.277  & \textbf{0.407 } & \textbf{0.271 } & 0.662  & 0.417  & \textbf{0.599 } & \textbf{0.380 } & 0.618  & 0.327  & \textbf{0.493 } & \textbf{0.309 } & 0.447  & 0.280  & \textbf{0.442 } & \textbf{0.273 } & 1.411  & 0.803  & \textbf{0.397 } & \textbf{0.254 } \\
 & {336} & 0.424  & 0.283  & \textbf{0.417 } & \textbf{0.279 } & 0.669  & 0.419  & \textbf{0.590 } & \textbf{0.374 } & 0.642  & 0.341  & \textbf{0.514 } & \textbf{0.320 } & \textbf{0.472 } & \textbf{0.289 } & 0.473  & 0.290  & 1.427  & 0.806  & \textbf{0.408 } & \textbf{0.261 } \\
 & {720} & 0.459  & 0.301  & \textbf{0.447 } & \textbf{0.293 } & 0.709  & 0.437  & \textbf{0.616 } & \textbf{0.385 } & 0.679  & \textbf{0.350 } & \textbf{0.568 } & 0.351  & \textbf{0.517 } & \textbf{0.307 } & 0.522  & 0.310  & \textbf{0.435 } & \textbf{0.298 } & 0.602  & 0.387 \\\cmidrule{2-22} 
& {Avg} & 0.422  & 0.282  & \textbf{0.416 } & \textbf{0.278 } & 0.688  & 0.427  & \textbf{0.613 } & \textbf{0.385 } & 0.633  & 0.333  & \textbf{0.511 } & \textbf{0.318 } & 0.466  & 0.286  & \textbf{0.463 } & \textbf{0.284 } & 0.912  & 0.544  & \textbf{0.444 } & \textbf{0.286 } \\\midrule
\multirow[c]{5}{*}{\rotatebox{90}{Weather}}  & {96} & 0.176  & 0.216  & \textbf{0.170 } & \textbf{0.208 } & 0.195  & \textbf{0.252 } & \textbf{0.187 } & 0.255  & 0.172  & 0.221  & \textbf{0.156 } & \textbf{0.204 } & 0.157  & 0.205  & \textbf{0.151 } & \textbf{0.201 } & 0.177  & 0.218  & \textbf{0.173 } & \textbf{0.210 }  \\
 & {192} & 0.225  & 0.257  & \textbf{0.223 } & \textbf{0.256 } & 0.239  & 0.299  & \textbf{0.228 } & \textbf{0.291 } & 0.220  & 0.260  & \textbf{0.207 } & \textbf{0.249 } & 0.204  & 0.248  & \textbf{0.202 } & \textbf{0.244 } & 0.223  & 0.258  & \textbf{0.216 } & \textbf{0.251 } \\
 & {336} & 0.281  & 0.299  & \textbf{0.279 } & \textbf{0.295 } & 0.282  & \textbf{0.333 } & \textbf{0.273 } & 0.335  & 0.280  & 0.302  & \textbf{0.265 } & \textbf{0.294 } & 0.264  & 0.293  & \textbf{0.264 }  & \textbf{0.292 } & \textbf{0.279 } & 0.299  & 0.280  & \textbf{0.291 } \\
 & {720} & 0.361  & 0.353  & \textbf{0.355 } & \textbf{0.347 } & \textbf{0.345 } & \textbf{0.381 } & 0.349  & 0.388  & 0.353  & 0.350  & \textbf{0.349 } & \textbf{0.343 } & 0.343  & \textbf{0.343 } & \textbf{0.341 } & 0.344  & \textbf{0.354 } & 0.349  & 0.356  & \textbf{0.343 } \\\cmidrule{2-22} 
& {Avg} & 0.260  & 0.281  & \textbf{0.256 } & \textbf{0.276 } & 0.265  & \textbf{0.316 } & \textbf{0.259 } & 0.317  & 0.256  & 0.283  & \textbf{0.244 } & \textbf{0.272 } & 0.242  & 0.272  & \textbf{0.239 } & \textbf{0.270 } & \textbf{0.258 } & 0.281  & 0.256  & \textbf{0.273 } \\
\bottomrule
\end{tabular}}
\caption{Full results of long-term forecasting. All the results are selected from 4 different prediction lengths $\{96, 192, 336, 720\}$, and the look-back length is fixed to 96 for all baselines. “Ori” denotes the original backbone. The better results are highlighted in \textbf{bold}.}
\label{full_results1}
\end{table*}

\begin{table*}[!t]
\resizebox{\textwidth}{!}{
\begin{tabular}{c|c|cc|cc|cc|cc|cc|cc|cc|cc|cc|cc}
    \toprule
\multicolumn{2}{c|}{\multirow{2}{*}{Methods}} & \multicolumn{4}{c|}{iTransformer} & \multicolumn{4}{c|}{DLinear} & \multicolumn{4}{c|}{TimesNet} & \multicolumn{4}{c|}{TimeXer} & \multicolumn{4}{c}{TimeBridge} \\
\cmidrule(lr){3-22}
 \multicolumn{2}{c|}{}& \multicolumn{2}{c|}{Ori} & \multicolumn{2}{c|}{+DPAD} & \multicolumn{2}{c|}{Ori} & \multicolumn{2}{c|}{+DPAD} & \multicolumn{2}{c|}{Ori} & \multicolumn{2}{c|}{+DPAD} & \multicolumn{2}{c|}{Ori} & \multicolumn{2}{c|}{+DPAD} & \multicolumn{2}{c|}{Ori} & \multicolumn{2}{c}{+DPAD} \\     \midrule
\multicolumn{2}{c|}{Metric} & MSE & MAE & MSE & MAE & MSE & MAE & MSE & MAE & MSE & MAE & MSE & MAE & MSE & MAE & MSE & MAE & MSE & MAE & MSE & MAE \\ \midrule
 \multirow[c]{5}{*}{\rotatebox{90}{PEMS03}}& {12} & 0.069  & 0.175  & \textbf{0.067 } & \textbf{0.169 } & 0.122  & 0.245  & \textbf{0.102 } & \textbf{0.218 } & 0.088  & 0.195  & \textbf{0.066 } & \textbf{0.172 } & \textbf{0.068 } & \textbf{0.179 } & 0.077  & 0.183  & 0.075  & 0.183  & \textbf{0.072 } & \textbf{0.178 } \\
 & {24} & \textbf{0.099 } & \textbf{0.210 } & 0.102  & 0.213  & 0.202  & 0.320  & \textbf{0.141 } & \textbf{0.257 } & 0.118  & 0.224  & \textbf{0.091 } & \textbf{0.198 } & \textbf{0.089 } & \textbf{0.204 } & 0.091  & 0.211  & 0.105  & 0.219  & \textbf{0.102 } & \textbf{0.212 } \\
 & {48} & \textbf{0.164 } & \textbf{0.275 } & 0.169  & 0.281  & 0.334  & 0.428  & \textbf{0.226 } & \textbf{0.329 } & 0.169  & 0.268  & \textbf{0.137 } & \textbf{0.244 } & 0.137  & 0.253  & \textbf{0.134 } & \textbf{0.249 } & 0.171  & 0.284  & \textbf{0.169 } & \textbf{0.278 }  \\
 & {96} & 0.711  & 0.651  & \textbf{0.417 } & \textbf{0.472 } & 0.459  & 0.517  & \textbf{0.319 } & \textbf{0.403 } & 0.239  & 0.330  & \textbf{0.206 } & \textbf{0.303 } & 0.427  & 0.483  & \textbf{0.236 } & \textbf{0.349 } & \textbf{0.262 } & \textbf{0.365 } & 0.277  & 0.370  \\\cmidrule{2-22} 
 & {Avg} & 0.260  & 0.327  & \textbf{0.188 } & \textbf{0.283 } & 0.279  & 0.377  & \textbf{0.197 } & \textbf{0.301 } & 0.153  & 0.254  & \textbf{0.125 } & \textbf{0.229 } & 0.180  & 0.279  & \textbf{0.134 } & \textbf{0.248 } & \textbf{0.153 } & 0.262  & 0.155  & \textbf{0.259 } \\\midrule
  \multirow[c]{5}{*}{\rotatebox{90}{PEMS04}}& {12} & 0.152 & 0.301  & \textbf{0.115 } & \textbf{0.241 } & 0.147  & 0.272  & \textbf{0.114 } & \textbf{0.233 } & 0.092  & 0.202  & \textbf{0.073 } & \textbf{0.179 } & 0.293  & 0.397  & \textbf{0.083 } & \textbf{0.198 } & 0.097  & 0.206  & \textbf{0.093 } & \textbf{0.198 } \\
 & {24} & 0.205 & 0.323  & \textbf{0.157 } & \textbf{0.280 } & 0.225  & 0.340  & \textbf{0.157 } & \textbf{0.275 } & 0.111  & 0.224  & \textbf{0.088 } & \textbf{0.198 } & 0.308  & 0.409  & \textbf{0.105 } & \textbf{0.229 } & \textbf{0.133 } & 0.243  & 0.134  & \textbf{0.240 }  \\
 & {48} & 0.265 & 0.370  & \textbf{0.229 } & \textbf{0.341 } & 0.356  & 0.437  & \textbf{0.239 } & \textbf{0.343 } & 0.152  & 0.266  & \textbf{0.118 } & \textbf{0.234 } & 0.339  & 0.425  & \textbf{0.146 } & \textbf{0.272 } & \textbf{0.215 } & 0.319  & 0.216  & \textbf{0.313 } \\
 & {96} & 0.451 & 0.483  & \textbf{0.293 } & \textbf{0.388 } & 0.453  & 0.505  & \textbf{0.312 } & \textbf{0.401 } & 0.197  & 0.308  & \textbf{0.167 } & \textbf{0.279 } & 0.367  & 0.441  & \textbf{0.219 } & \textbf{0.343 } & 0.329  & 0.412  & \textbf{0.324 } & \textbf{0.407 } \\\cmidrule{2-22} 
 & {Avg} & 0.268  & 0.369  & \textbf{0.199 } & \textbf{0.312 } & 0.295  & 0.388  & \textbf{0.205 } & \textbf{0.313 } & 0.138  & 0.250  & \textbf{0.111 } & \textbf{0.222 } & 0.326  & 0.418  & \textbf{0.138 } & \textbf{0.261 } & 0.193  & 0.295  & \textbf{0.192 } & \textbf{0.289 }  \\\midrule
  \multirow[c]{5}{*}{\rotatebox{90}{PEMS07}}& {12} & 0.068  & 0.169  & \textbf{0.065 } & \textbf{0.161 } & 0.116  & 0.241  & \textbf{0.100 } & \textbf{0.210 } & 0.075  & 0.179  & \textbf{0.063 } & \textbf{0.162 } & 0.061  & 0.165  & \textbf{0.056 } & \textbf{0.161 } & 0.067  & 0.168  & \textbf{0.066 } & \textbf{0.163 } \\
 & {24} & 0.087  & 0.190  & \textbf{0.081 } & \textbf{0.187 } & 0.209  & 0.327  & \textbf{0.149 } & \textbf{0.261 } & 0.083  & 0.198  & \textbf{0.080 } & \textbf{0.183 } & 0.071  & 0.177  & \textbf{0.067 } & \textbf{0.171 } & \textbf{0.095 } & 0.200  & 0.096  & \textbf{0.196 } \\
 & {48} & 0.122  & 0.231  & \textbf{0.118 } & \textbf{0.225 } & 0.397  & 0.456  & \textbf{0.269 } & \textbf{0.347 } & 0.128  & 0.235  & \textbf{0.118 } & \textbf{0.221 } & \textbf{0.100 } & \textbf{0.208 } & 0.109  & 0.211  & \textbf{0.145 } & 0.252  & 0.148  & \textbf{0.248 } \\
 & {96} & 0.159  & 0.267  & \textbf{0.152 } & \textbf{0.258 } & 0.592  & 0.552  & \textbf{0.374 } & \textbf{0.422 } & \textbf{0.150 } & \textbf{0.253 } & 0.162  & 0.261  & \textbf{0.120 } & \textbf{0.221 } & 0.124  & 0.225  & \textbf{0.206 } & \textbf{0.310 } & 0.211  & 0.312 \\\cmidrule{2-22} 
 & {Avg} & 0.109  & 0.214  & \textbf{0.104 } & \textbf{0.207 } & 0.328  & 0.394  & \textbf{0.223 } & \textbf{0.310 } & 0.109  & 0.216  & \textbf{0.105 } & \textbf{0.206 } & \textbf{0.088 } & 0.192  & 0.089  & \textbf{0.192 }  & \textbf{0.128 } & 0.232  & 0.130  & \textbf{0.230 } \\\midrule
  \multirow[c]{5}{*}{\rotatebox{90}{PEMS08}}& {12} & 0.081  & 0.183  & \textbf{0.078 } & \textbf{0.179 } & 0.153  & 0.259  & \textbf{0.148 } & \textbf{0.234 } & 0.158  & 0.192  & \textbf{0.082 } & \textbf{0.185 } & 0.146  & \textbf{0.198 } & \textbf{0.088 } & 0.204  & 0.080  & 0.184  & \textbf{0.075 } & \textbf{0.178 } \\
 & {24} & 0.118  & 0.222  & \textbf{0.113 } & \textbf{0.219 } & 0.238  & 0.357  & \textbf{0.192 } & \textbf{0.277 } & 0.112  & 0.219  & \textbf{0.108 } & \textbf{0.212 } & 0.171  & \textbf{0.221 } & \textbf{0.125 } & 0.241  & 0.112  & 0.217  & \textbf{0.110 } & \textbf{0.213 } \\
 & {48} & \textbf{0.202 } & \textbf{0.292 } & 0.210  & 0.294  & 0.473  & 0.515  & \textbf{0.310 } & \textbf{0.353 } & 0.231  & \textbf{0.198 } & \textbf{0.166 } & 0.221  & 0.220  & \textbf{0.270 } & \textbf{0.208 } & 0.283  & 0.186  & 0.278  & \textbf{0.181 } & \textbf{0.272 } \\
 & {96} & 0.395  & 0.415  & \textbf{0.294 } & \textbf{0.326 } & 0.748  & 0.646  & \textbf{0.441 } & \textbf{0.429 } & 0.291  & 0.338  & \textbf{0.287 } & \textbf{0.324 } & \textbf{0.285 } & \textbf{0.303 } & 0.296  & 0.321  & 0.296  & 0.344 & \textbf{0.292 } & \textbf{0.340 } \\\cmidrule{2-22} 
 & {Avg} & 0.199  & 0.278  & \textbf{0.173 } & \textbf{0.254 } & 0.403  & 0.444  & \textbf{0.272 } & \textbf{0.323 } & 0.198  & 0.236  & \textbf{0.160 } & \textbf{0.235 } & 0.205  & \textbf{0.248 } & \textbf{0.179 } & 0.262  & 0.168  & 0.255  & \textbf{0.164 } & \textbf{0.251 } \\
\bottomrule
\end{tabular}}
\caption{Full results of short-term forecasting. All the results are selected from 4 different prediction lengths $\{12, 24, 48, 96\}$, and the look-back length is fixed to 96 for all baselines. The better results are highlighted in \textbf{bold}.}
\label{full_results2}
\end{table*}

\begin{table*}[!t]
    \resizebox{\textwidth}{!}{
	\begin{tabular}{c|c|cc|cc|cc|cc|cc|cc|cc|cc|cc|cc}
		\toprule
		\multicolumn{2}{c|}{Case} & \multicolumn{2}{c|}{\textbf{+DPAD}} & \multicolumn{2}{c|}{\ding{172}w/o DDP} & \multicolumn{2}{c|}{\ding{173}w/o Common} & \multicolumn{2}{c|}{\ding{174}w/o Rare} & \multicolumn{2}{c|}{\ding{175}w/ Additive} & \multicolumn{2}{c|}{\ding{176}w/ Mean} & \multicolumn{2}{c|}{\ding{177}w/o $\mathcal{L}_{\text{DGL}}$} & \multicolumn{2}{c|}{\ding{178}w/o $\mathcal{L}_{\text{sep}}$} & \multicolumn{2}{c|}{\ding{179}w/o $\mathcal{L}_{\text{rare}}$} & \multicolumn{2}{c}{\ding{180}w/o $\mathcal{L}_{\text{div}}$}\\
		\cmidrule(lr){1-2} \cmidrule(lr){3-4} \cmidrule(lr){5-6} \cmidrule(lr){7-8} \cmidrule(lr){9-10} \cmidrule(lr){11-12} \cmidrule(lr){13-14} \cmidrule(lr){15-16} \cmidrule(lr){17-18} \cmidrule(lr){19-20} \cmidrule(lr){21-22}
		\multicolumn{2}{c|}{Metric} & MSE & MAE & MSE & MAE & MSE & MAE & MSE & MAE & MSE & MAE & MSE & MAE & MSE & MAE & MSE & MAE & MSE & MAE & MSE & MAE\\
		\midrule
		\multirow{5}{*}{\rotatebox{90}{Electricity}} & 96    & \textbf{0.147} & \textbf{0.234} & 0.148 & 0.241 & 0.152  & 0.243  & 0.155  & 0.244  & 0.153  & 0.242  & 0.149  & 0.245  & 0.151  & 0.240  & 0.153  & 0.242  & 0.154  & 0.239  & 0.151  & 0.240  \\
          & 192   & \textbf{0.162} & 0.251 & 0.167 & \textbf{0.248} & 0.167  & 0.259  & 0.170  & 0.261  & 0.169  & 0.260  & 0.167  & 0.263  & 0.171  & 0.264  & 0.168  & 0.262  & 0.166  & 0.259  & 0.165  & 0.260  \\
          & 336   & \textbf{0.172} & \textbf{0.271} & 0.179 & 0.271 & 0.178  & 0.271  & 0.181  & 0.273  & 0.182  & 0.272  & 0.181  & 0.275  & 0.183  & 0.275  & 0.180  & 0.273  & 0.182  & 0.275  & 0.184  & 0.277  \\
          & 720   & \textbf{0.201} & \textbf{0.289} & 0.208 & 0.298 & 0.216  & 0.303  & 0.225  & 0.310  & 0.213  & 0.298  & 0.212  & 0.294  & 0.217  & 0.305  & 0.209  & 0.301  & 0.212  & 0.297  & 0.208  & 0.299  \\
		\cmidrule(lr){2-22}
		& Avg & \textbf{0.170} & \textbf{0.261} & 0.175 & 0.264 & 0.178  & 0.269  & 0.183  & 0.272  & 0.179  & 0.268  & 0.177  & 0.269  & 0.181  & 0.271  & 0.178  & 0.269  & 0.179  & 0.268  & 0.177  & 0.269\\
		\midrule
		\multirow{5}{*}{\rotatebox{90}{Weather}} & 96    & \textbf{0.17} & \textbf{0.208} & 0.176 & 0.216 & 0.175  & 0.214  & 0.175  & 0.215  & 0.178  & 0.216  & 0.173  & 0.212  & 0.177  & 0.216  & 0.173  & 0.212  & 0.175  & 0.216  & 0.172  & 0.212  \\
          & 192   & \textbf{0.223} & \textbf{0.256} & 0.225 & 0.257 & 0.229  & 0.261  & 0.230  & 0.262  & 0.228  & 0.259  & 0.227  & 0.260  & 0.231  & 0.263  & 0.229  & 0.260  & 0.229  & 0.259  & 0.230  & 0.261  \\
          & 336   & \textbf{0.279} & \textbf{0.295} & 0.281 & 0.299 & 0.283  & 0.300  & 0.285  & 0.302  & 0.283  & 0.299  & 0.282  & 0.301  & 0.286  & 0.302  & 0.284  & 0.300  & 0.283  & 0.303  & 0.282  & 0.301  \\
          & 720   & \textbf{0.355} & \textbf{0.347} & 0.361 & 0.353 & 0.362  & 0.353  & 0.361  & 0.351  & 0.361  & 0.352  & 0.359  & 0.352  & 0.363  & 0.352  & 0.360  & 0.352  & 0.359  & 0.352  & 0.358  & 0.352  \\
		\cmidrule(lr){2-22}
		& Avg   & \textbf{0.256} & \textbf{0.276} & 0.260  & 0.281 & 0.262  & 0.282  & 0.263  & 0.282  & 0.262  & 0.282  & 0.260  & 0.281  & 0.264  & 0.283  & 0.262  & 0.281  & 0.262  & 0.283  & 0.261  & 0.282  \\
		\midrule
		\multirow{5}{*}{\rotatebox{90}{Traffic}} & 96    & \textbf{0.392 } & \textbf{0.268 } & 0.393 & 0.269 & 0.477  & 0.337  & 0.395  & 0.272  & 0.396  & 0.270  & 0.459  & 0.325  & 0.457  & 0.323  & 0.426  & 0.314  & 0.433  & 0.310  & 0.468  & 0.333  \\
          & 192   & \textbf{0.407} & \textbf{0.271} & 0.412 & 0.277 & 0.414  & 0.277  & 0.413  & 0.277  & 0.412  & 0.276  & 0.413  & 0.279  & 0.418  & 0.282  & 0.478  & 0.333  & 0.415  & 0.280  & 0.412  & 0.277  \\
          & 336   & \textbf{0.417 } & \textbf{0.279} & 0.424 & 0.283 & 0.425  & 0.283  & 0.427  & 0.284  & 0.425  & 0.284  & 0.425  & 0.286  & 0.429  & 0.286  & 0.424  & 0.282  & 0.493  & 0.340  & 0.425  & 0.282  \\
          & 720   & \textbf{0.447} & \textbf{0.293} & 0.459 & 0.301 & 0.531  & 0.359  & 0.457  & 0.300  & 0.454  & 0.299  & 0.457  & 0.301  & 0.459  & 0.302  & 0.529  & 0.357  & 0.461  & 0.301  & 0.458  & 0.303  \\
		\cmidrule(lr){2-22}
		& Avg   & \textbf{0.416 } & \textbf{0.278 } & 0.422 & 0.282 & 0.462  & 0.314  & 0.423  & 0.283  & 0.422  & 0.282  & 0.439  & 0.298  & 0.441  & 0.298  & 0.464  & 0.322  & 0.450  & 0.308  & 0.441  & 0.299  \\
		\midrule
		\multirow{5}{*}{\rotatebox{90}{Solar}} & 96    & \textbf{0.199} & \textbf{0.221} & 0.207 & 0.237 & 0.206  & 0.239  & 0.204  & 0.233  & 0.205  & 0.247  & 0.203  & 0.236  & 0.210  & 0.241  & 0.208  & 0.241  & 0.209  & 0.240  & 0.206  & 0.243  \\
          & 192   & \textbf{0.237} & \textbf{0.254} & 0.242 & 0.264 & 0.244  & 0.268  & 0.240  & 0.260  & 0.239  & 0.267  & 0.239  & 0.261  & 0.240  & 0.266  & 0.239  & 0.265  & 0.240  & 0.266  & 0.241  & 0.259  \\
          & 336   & \textbf{0.249} & \textbf{0.273} & 0.251 & 0.277 & 0.254  & 0.278  & 0.252  & 0.275  & 0.253  & 0.278  & 0.250  & 0.275  & 0.255  & 0.281  & 0.254  & 0.280  & 0.255  & 0.279  & 0.253  & 0.280  \\
          & 720   & \textbf{0.250} & \textbf{0.276} & 0.251 & 0.278 & 0.256  & 0.280  & 0.252  & 0.278  & 0.253  & 0.279  & 0.254  & 0.277  & 0.260  & 0.282  & 0.254  & 0.279  & 0.254  & 0.280  & 0.253  & 0.279  \\
		\cmidrule(lr){2-22}
		& Avg   & \textbf{0.233} & \textbf{0.256} & 0.237 & 0.262 & 0.240  & 0.266  & 0.237  & 0.262  & 0.237  & 0.268  & 0.237  & 0.262  & 0.241  & 0.267  & 0.239  & 0.266  & 0.239  & 0.266  & 0.238  & 0.265  \\
		\bottomrule
	\end{tabular}
}
\caption{Full Results of ablation study. All the results are selected from 4 different prediction lengths \{96, 192, 336, 720\}, and the look-back length is fixed to 96 for all baselines. The better results are highlighted in \textbf{bold}.}
\label{full_ablation}
\end{table*}

\end{document}